\documentclass{vgtc}

\graphicspath{{figures/}{pictures/}{images/}{./}}

\usepackage{times}

\usepackage{tabu}
\usepackage{booktabs}
\usepackage{tabularx}
\usepackage{adjustbox}
\usepackage{mathptmx}
\usepackage{caption,subcaption}
\usepackage{amsmath}
\usepackage{xcolor}
\usepackage[normalem]{ulem}
\usepackage{algorithm}
\usepackage{algpseudocode}

\onlineid{1772}
\vgtccategory{Research}
\vgtcinsertpkg

\title{RoomRecon: High-Quality Textured Room Layout Reconstruction on Mobile Devices}

\author{
  Seok Joon Kim\thanks{Affiliated with MAXST Co., Ltd.} \\
  \scriptsize MAXST Co., Ltd.
\and
  Dinh Duc Cao\footnotemark[1]\thanks{Equal contribution.} \\
  \scriptsize MAXST Co., Ltd.
\and
  Federica Spinola\footnotemark[2]\thanks{e-mail: spinolafederica@yahoo.com.} \\
  \scriptsize Independent Researcher
\and
  Se Jin Lee\footnotemark[1] \\
  \scriptsize MAXST Co., Ltd.
\and
  Kyu Sung Cho\footnotemark[1] \\
  \scriptsize MAXST Co., Ltd.
}

\teaser{
  \centering
  \includegraphics[width=1.0\linewidth,height=0.25\linewidth]{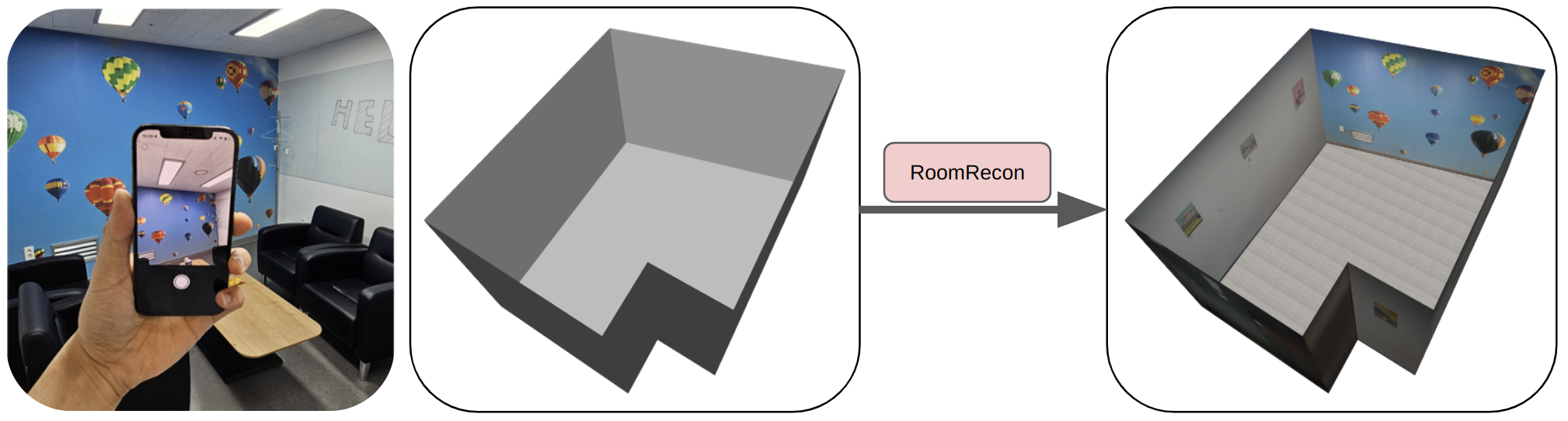}
  \caption{\textbf{Creating realistic textured 3D room models on mobile devices}: Capitalizing on the wide availability of mobile devices, RoomRecon creates textured 3D replicas of indoor spaces, readily applicable for use in VR and AR applications.}
  \label{fig:teaser}
}

\abstract{
Widespread RGB-Depth (RGB-D) sensors and advanced 3D reconstruction technologies facilitate the capture of indoor spaces, improving the fields of augmented reality (AR), virtual reality (VR), and extended reality (XR). Nevertheless, current technologies still face limitations, such as the inability to reflect minor scene changes without a complete recapture, the lack of semantic scene understanding, and various texturing challenges that affect the 3D model's visual quality. These issues affect the realism required for VR experiences and other applications such as in interior design and real estate. To address these challenges, we introduce RoomRecon, an interactive, real-time scanning and texturing pipeline for 3D room models. We propose a two-phase texturing pipeline that integrates AR-guided image capturing for texturing and generative AI models to improve texturing quality and provide better replicas of indoor spaces. Moreover, we suggest to focus only on permanent room elements such as walls, floors, and ceilings, to allow for easily customizable 3D models. We conduct experiments in a variety of indoor spaces to assess the texturing quality and speed of our method. The quantitative results and user study demonstrate that RoomRecon surpasses state-of-the-art methods in terms of texturing quality and on-device computation time.
}

\keywords{Room layout, Indoor 3D reconstruction, Texturing, AR-assisted image capturing, Mobile application.}

\begin{document}

\maketitle

\section{Introduction}
\label{Introduction}
Widespread RGB-Depth (RGB-D) sensors and improved 3D reconstruction methods~\cite{BundleFusion, KinectFusion, ElasticFusion} allow to better capture indoor spaces, easing the creation of augmented reality (AR), virtual reality (VR), and extended reality (XR) applications~\cite{review1}. The recent integration of these 3D reconstruction technologies into mobile devices allows individuals to quickly and effortlessly capture their own spaces~\cite{ARKIT, ARCore, MobileRecon}. In fact, accurate replicas of scenes are sought after in fields like interior design and architecture~\cite{interior2, interior1}, where precise representations facilitate visualization, planning, and effective communication between coworkers. Additionally, such replicas offer benefits in real estate, enabling virtual property tours for potential buyers.\\
However, despite these technological advancements, several limitations persist in constructing 3D models. Firstly, the captured and reconstructed scenes are static, requiring users to fully recapture them when internal changes occur such as furniture rearrangements or new wall decorations. Secondly, current algorithms lack a semantic understanding of the scene, preventing selective geometry alterations. For instance, users cannot decide to remove the object \textit{table} from their room's 3D model, nor can they choose to keep the object \textit{bed}. Thirdly, environmental and technological challenges disrupt the texturing process, which is crucial for enhancing the realism of 3D models and improving user experience. The environmental constraints, such as obstructions by furniture, lead to untextured areas in the 3D models. Technological limitations, including blurred images and camera pose drift, produce textured outputs with blurs, seams, and lighting inconsistencies that disrupt visual coherence. These discrepancies are common problems of state-of-art (SOTA) texturing methods~\cite{PatchMatch, SEAMLESS, MVSTexturing}. All of these issues affect the usability of 3D models in VR or AR applications, which require realistic, complete and dynamic virtual environments for immersive user experiences. \\
With the growing popularity of diffusion models, several noteworthy methods such as Text2Room~\cite{Text2Room}, ControlRoom3D~\cite{ControlRoom3D} and RoomDreamer~\cite{RoomDreamer} can produce aesthetically pleasing indoor spaces. From fully generated to more controlled 3D models, these methods can create many environments suitable for VR applications. Nevertheless, none of these methods tackle the problem of creating accurate virtual replicas of indoor spaces. \\
To address the challenges of accurately replicating indoor spaces with high realism and completeness, we introduce RoomRecon, an interactive, real-time 3D reconstruction and texturing mobile application for 3D room modeling. Firstly, RoomRecon proposes to capture only permanent room elements such as walls, floors, and ceilings (i.e. the room's layout). This reduces the need for recaptures as soon as furniture changes, allowing for durable and customizable representations of indoor spaces. Whilst it is beyond the scope of this work, we note that dynamic objects can be added to our 3D models by using existing CAD model databases~\cite{scene2Cad, vid2cad} and pre-captured textured meshes~\cite{3DFuture}. Secondly, RoomRecon proposes to selectively capture a small number of high-quality images to be used in the texturing phase, through a dedicated user interface (UI) and AR cues during the scanning process. This aims to improve texturing quality by ensuring that surfaces are textured by as few images as possible, helping to reduce blur, seams and structural inconsistencies. By addressing the texturing's environmental and technological challenges, RoomRecon aims to model 3D scenes that are accurate replicas of indoor spaces, distinguishing our method from previous solutions~\cite{Text2Room, ControlRoom3D, RoomDreamer}.\\
To summarise, our contributions are as follows:
\begin{enumerate}
\vspace{-7pt}
    \item We present RoomRecon, a mobile framework that efficiently generates complete and high-fidelity 3D models of rooms.
    \vspace{-7pt}
    \item We present a divide-and-conquer strategy for surface texturing, relying on AR-guided human involvement.
    \vspace{-7pt}
    \item We introduce the Plane2Image rendering module, which enables editing textured surfaces in 2D image space by using common image processing techniques or more advanced diffusion-based generative models.
    \vspace{-7pt}
    \item We conduct extensive experiments and user studies on a variety of indoor spaces, achieving superior results both quantitatively and qualitatively compared to SOTA texturing methods.
    \vspace{-7pt}
\end{enumerate}
\vspace{-11pt}

\begin{figure*}[ht]
  \centering
  \includegraphics[width=1.0\linewidth]{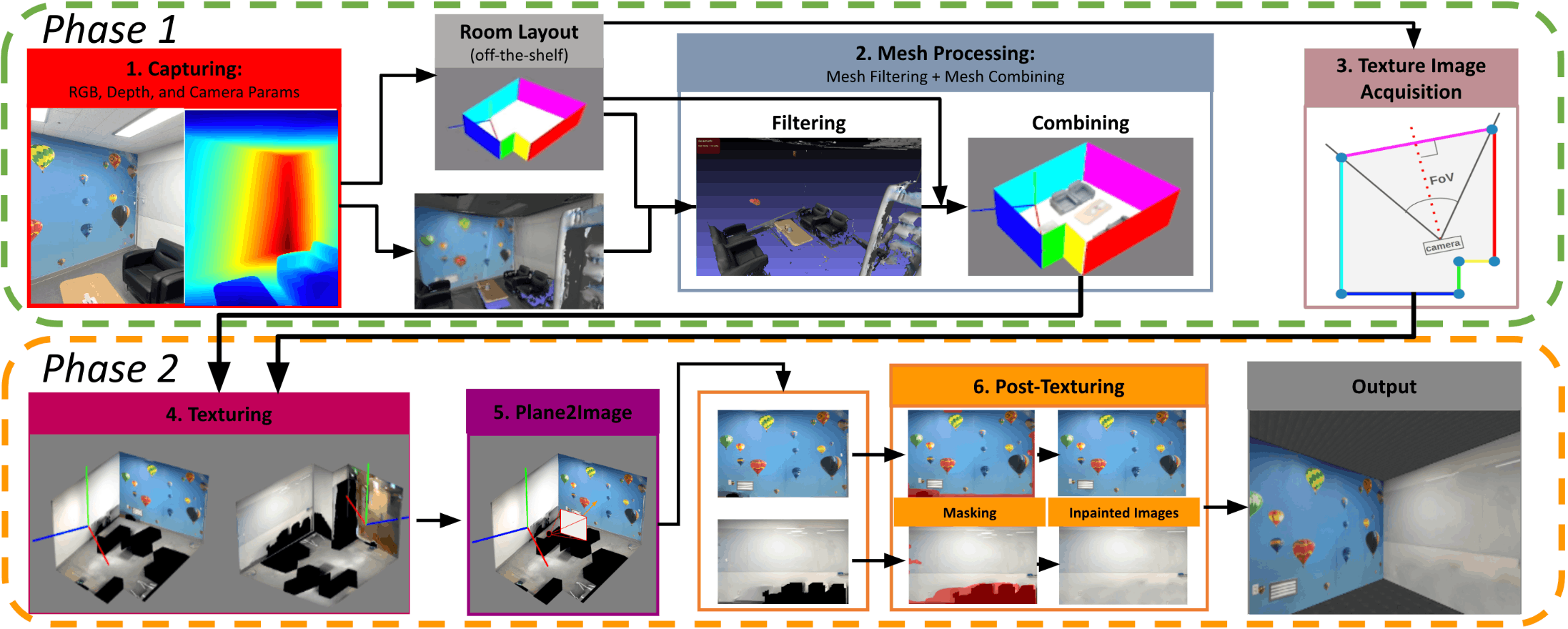}
  \caption{
    \textbf{Method overview}: Our system constructs detailed textured 3D room models through a two-phase pipeline. Initially, the \textbf{Capturing step} collects a calibrated set of RGB-D images and creates the original mesh ($M_{\text{original}}$). In the \textbf{Mesh Processing step}, this mesh is filtered using the room layout to produce a filtered mesh ($M_{\text{filtered}}$), which is then integrated with the layout to create the combined mesh ($M_{\text{combined}}$). A new set of RGB images is captured in the \textbf{Texture Image Acquisition step}, which is utilized in the second phase's \textbf{Texturing step}. Here, the $M_{\text{combined}}$ and these RGB images are used to apply textures to the mesh. Following this, the \textbf{Plane2Image step} generates a set of images corresponding to the room's walls. These images are then refined in the \textbf{Post-Texturing step}, which finalizes the pipeline.
  }
  \vspace{-15pt}
  \label{fig:overview}
\end{figure*}

\section{Background and Related Work}

\subsection{Texturing for 3D Reconstruction}
Texturing aims to ensure sharp and realistic appearances of 3D models to enhance user experience.
As Zhao et al.~\cite{advertinit} outlines, texturing methods can be categorized into four main types: average-based, warping-based, learning-based, and assignment-based. The most relevant for our work is the assignment-based method, which assigns specific images to each face of the model using a Markov Random Field (MRF) based optimization. Assignment-based methods utilize patches of original high-resolution images, leading to very sharp texturing results. Building on this foundation, Waechter et al.~\cite{MVSTexturing}, Fu et al.~\cite{SEAMLESS, TextureMapping} and Zhao et al.~\cite{advertinit}
apply pairwise costs in the MRF optimization to enforce the texturing of neighboring faces with the same image, ensuring local consistency and sharpness. Taking the work of Waechter et al.~\cite{MVSTexturing} further, Fu et al.~\cite{SEAMLESS} introduce detail-aware texture chart partitions to avoid creating partitions in areas of the mesh with high detail, thus preserving texture integrity in these detailed regions.
In contrast, average-based approaches often produce blurry textures or ghosting effects due to the process of averaging color values from multiple images. Zhou et al.~\cite{COLORMAPOPTIM} assign colors to mesh vertices by averaging the colors from all images capturing each vertex, a technique highly dependent on the mesh's resolution and generally requiring a very fine mesh to achieve satisfactory quality.
Wang et al.~\cite{PlaneOpt} propose using 3D texels positioned within mesh faces.
In the domain of learning-based texturing, Zhao et al.~\cite{advertinit} and Huang et al.~\cite{adversarial} leverage generative adversarial networks (GANs)~\cite{GAN} that are trained to differentiate between original and rendered images, thereby refining the texturing process. Finally, the work of Zhou et al.~\cite{COLORMAPOPTIM} and Fu et al.~\cite{TextureMapping} stand out for their use of warping-based methods to achieve more accurate texture mappings. Given the hardware limitations of mobile devices and the benefits in speed and textured output clarity, we have adopted assignment-based texturing methods~\cite{SEAMLESS, MVSTexturing} as the foundation for our approach.

\subsection{Image Inpainting}
Image inpainting is a process that repairs, erases, and fills data within masked areas of an image to restore or create content that blends with surrounding areas, rendering changes imperceptible to casual observers~\cite{inpaintingReview}. Whilst simple methods relying on image interpolation~\cite{interpolInpainting} can provide adequate results in some cases, recent learning-based methods \cite{ZIT, MAT, LamaInpainting} show better performance across a variety of applications. LAMA~\cite{LamaInpainting} proposes an inpainting network based on Fast Fourier Convolutions (FFCs) that create large receptive field, capturing the global context of images. MAT~\cite{MAT} presents a novel transformer-based model for large hole inpainting in high-resolution images. Attention mechanisms can learn better long range dependencies for structure recovery, but are slow, especially for large image sizes. ZIT~\cite{ZIT} addresses these issues by proposing an additional structure restorer that facilitates incremental image inpainting.
The recent development of diffusion models provides the ability to generate detailed images from text prompts~\cite{Photoshop, DALLE, LDM, Zhuang2023ATI}. Its applications extend to many generative functions including image inpainting. In this paper, we employ ZIT~\cite{ZIT} to correct texturing defects and fill untextured regions.

\section{System Design}

\subsection{Problem Statement}
\label{SystemDesign:problemStatement}
Our work aims to achieve a complete and precise textured room layout that closely replicates the appearance of rooms. However, the texturing process faces complications inherent to capturing indoor spaces and stemming from the texturing methods used. Environmental challenges include occlusions by furniture that result in untextured areas, shadows cast by objects affecting wall textures, and difficulties in accurately capturing thin objects (e.g.: clothes hangers, chair legs), often leading to incomplete meshes. These problems are particularly present for floor areas and lower regions of the wall, affecting their texture.
Ceilings pose texturing challenges due to uncomfortable image capture positions and limited mobility caused by furniture, resulting in images often taken at slanted angles. This can lead to textures with lower quality due to distortions and misalignments.
Additionally, the repetitive patterns on ceilings and floors, characterized by numerous lines between samples, complicate the texturing process. In fact, even slight deviations in camera pose can lead to misaligned lines, resulting in noticeable errors in the final textured output. Finally, texturing transparent or reflective surfaces, such as glass walls and mirrors, can also be challenging. Using multiple images to texture such planar structures often results in viewpoint inconsistencies, as each image captures a different external scene and undesired reflections. \\
Technological problems associated with texturing algorithms are mostly observed in assignment-based methods such as MVSTex~\cite{MVSTexturing} and SeamlessTex~\cite{SEAMLESS}. While these methods can generate photorealistic textures from high-resolution images, they are prone to artifacts stemming from geometric inaccuracies and camera pose drift. These issues become particularly visible when multiple images are used to texture a single wall, often resulting in noticeable seams. For more details regarding environmental and technological issues, refer to the supplementary materials (Section A.4).

\subsection{Overview}
\label{SystemDesign:overview}
Our system aims to reconstruct rooms with high realism through a two-phase pipeline, as shown in \cref{fig:overview}. The first phase involves capturing an indoor space using an RGB-D sensor, such as an iPhone, to collect RGB images, depth maps, camera parameters, and the original mesh (\(M_{\text{original}}\)), described in \cref{SystemDesign:Capturing}. A standard algorithm then generates a 3D room layout, detailing plane normals, wall corners, and floor and ceiling heights, as referenced in various studies \cite{AppleRoomPlan, scene2Cad, Pq-transformer, OMNISUPERVISED}. This layout, along with \(M_{\text{original}}\), is processed to generate a composite mesh (\cref{SystemDesign:MeshProcessing}). The Texture Images Acquisition module captures a small set of high-quality images for texturing (\cref{SystemDesign:TextureImageAcquAndTexturing}). The second phase starts with the Texturing step, producing a fully textured 3D model (\cref{SystemDesign:Texturing}). Each textured plane is converted into an image via the Plane2Image module (\cref{SystemDesign:Plane2Image}), followed by texture refinement in the Post-Texturing stage, which includes on-device and cloud-based enhancements (\cref{SystemDesign:PostTexturing}). Ultimately, our pipeline delivers precise 3D replicas of indoor spaces.

\subsection{Capturing}
\label{SystemDesign:Capturing}

The primary objective of the \textbf{Capturing step} is to scan a room, resulting in a collection of color and depth images along with their corresponding camera parameters (intrinsics and extrinsics). In the remainder of this paper, we will refer to these images as $C_{\text{original}}$ and $D_{\text{original}}$, respectively. From the ensemble of camera intrinsic parameters (focal lengths$f_x$ and $f_y$, and optical centers $c_x$ and $c_y$) we derive $K_{\text{average}}$. $K_{\text{average}}$ represents the mean intrinsic matrix, calculated by averaging the parameters of all cameras involved in the scan.
To improve the capturing process, we develop a UI, depicted in \cref{fig:method:captureView}, which displays real-time visualizations of scanned and unscanned areas, as a toggle option. This feature helps in scanning all areas, ensuring complete meshes. Quality meshes are crucial, especially for occlusion tests to be accurately performed in the Texturing phase in \cref{SystemDesign:Texturing}, which directly affect texturing quality. We refer the reader to the supplementary (Section A.4) for more information about the occlusion tests.\\
Our UI relies on the creation of the room's mesh on-the-fly. The mesh formation algorithm is based on Newcombe's work, KinectFusion~\cite{KinectFusion}. We exclude its pose tracking module, and instead we leverage $D_{original}$ and the camera parameters from Apple Inc.~\cite{ARKIT} for the depth fusion process. The resulting mesh is referred to as $M_{\text{original}}$ and serves as input to the Mesh Processing step in \cref{SystemDesign:MeshProcessing}.

\subsection{Mesh Processing}
\label{SystemDesign:MeshProcessing}
\begin{figure}[ht!]
\centering
\begin{subfigure}{0.67\columnwidth}
\centering
  \includegraphics[width=0.7\linewidth]{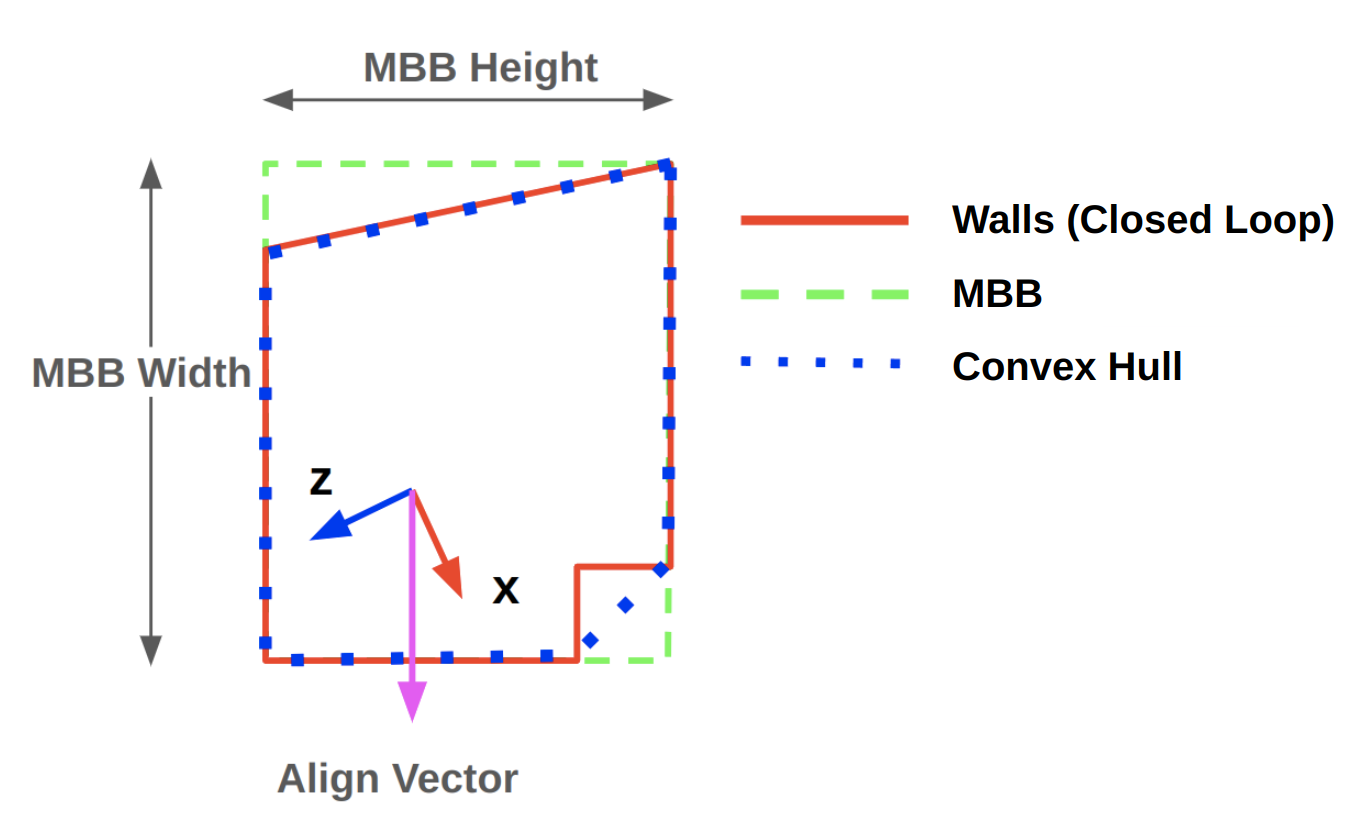}
  \caption{Minimum bounding box (MBB) calculation.}
  \label{fig:mbbAndCDT}
\end{subfigure}%
\hfill
\begin{subfigure}{0.33\columnwidth}
\centering
  \includegraphics[width=0.7\linewidth]{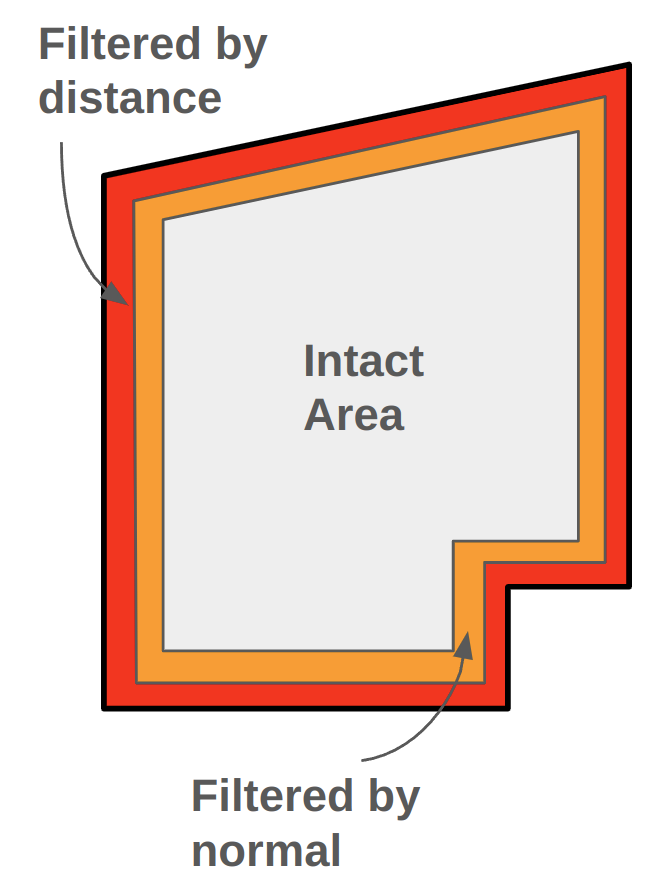}
  \caption{Mesh filtering scheme.}
  \label{fig:filteringScheme}
\end{subfigure}%
\hfill
\begin{subfigure}{\columnwidth}
\centering
  \includegraphics[width=0.8\linewidth]{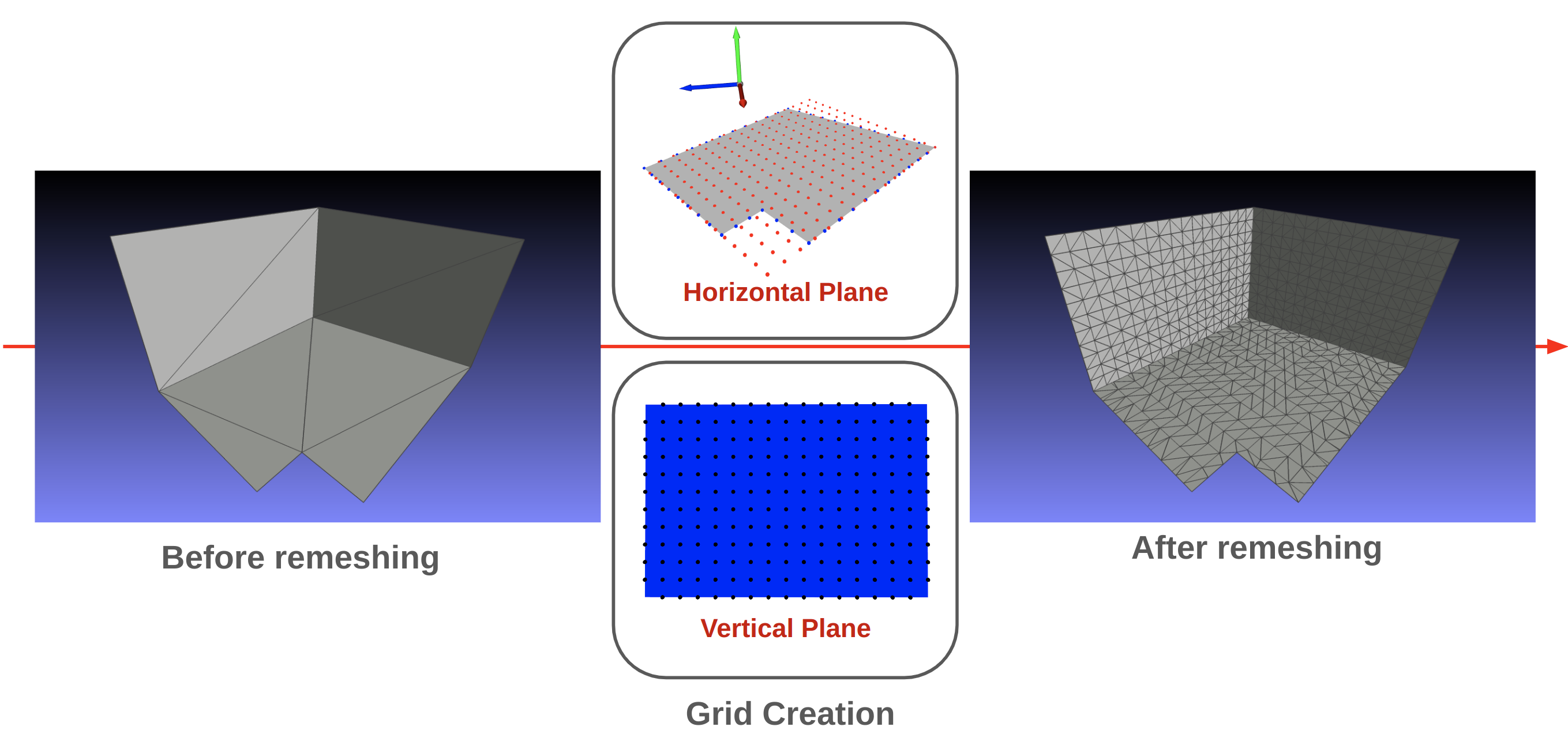}
  \caption{Remeshing scheme for structures of the room layout.}
  \label{fig:remeshingScheme}
\end{subfigure}
\hfill
\begin{subfigure}{0.48\columnwidth}
\centering
  \includegraphics[width=0.8\linewidth]{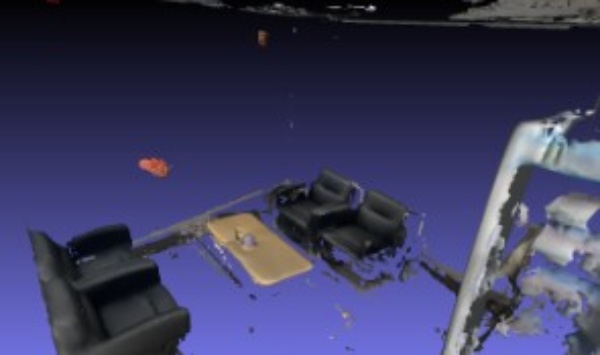}
  \caption{Filtered mesh $M_{\text{filtered}}$.}
  \label{fig:filteredMesh}
\end{subfigure}%
\hfill
\begin{subfigure}{0.48\columnwidth}
\centering
  \includegraphics[width=0.8\linewidth]{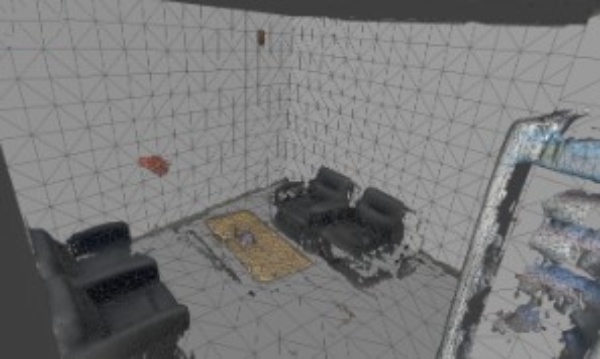}
  \caption{Combined mesh $M_{\text{combined}}$.}
  \label{fig:combinedMesh}
\end{subfigure}%
\vspace{-7pt}
\caption{Processes and outputs of the Mesh Processing step.}
\vspace{-7pt}
\label{fig:meshProcessingMethods}
\end{figure}

Our system categorizes rooms into open rooms and closed rooms. A closed room is defined by vertical walls forming a loop, enabling the creation of a watertight mesh fully enclosed by walls. In contrast, an open room lacks a closed loop of walls. For both room types, our pipeline initiates the \textbf{Mesh Processing step} by calculating a 2D Minimum Bounding Box (MBB)~\cite{MBB} in the X-Z coordinate plane, based on the previously computed room layout (see \cref{fig:mbbAndCDT} for the coordinate system). The MBB represents the smallest 2D rectangular box that entirely encloses all vertical walls of a room. As illustrated in \cref{fig:mbbAndCDT}, the outputs of the MBB calculation are $\text{MBB}_{\text{width}}$, $\text{MBB}_{\text{height}}$ and $\text{MBB}_{\text{align}}$. $\text{MBB}_{\text{width}}$ and $\text{MBB}_{\text{height}}$ define the spatial dimensions of the room. $\text{MBB}_{\text{align}}$, or Align Vector, is aligned with the $\text{MBB}_{\text{width}}$ edge, providing a reference for the subsequent processing stages in \cref{SystemDesign:Plane2Image} and \cref{SystemDesign:PostTexturing}. Subsequently, the original mesh, $M_{\text{original}}$, is filtered using the room layout. The filtering step focuses on meshes excluding the room layout's mesh. Its goal is to eliminate meshes too close to the walls. The mesh must be filtered sufficiently for successful texturing of the layout's planar structures, while retaining enough information about object meshes for occlusion tests. This process, illustrated in \cref{fig:filteringScheme}, is based on two criteria: (1) \textbf{Loop-Based}: This method eliminates mesh faces outside the 2D polygonal loop, defined by walls and the convex hull (see \cref{fig:mbbAndCDT}) for closed rooms and open rooms respectively. (2) \textbf{Distance and Normal-Based}: Firstly, all mesh faces close to the walls are removed, shown as \textit{Filtered by Distance} in \cref{fig:filteringScheme}. Then, further away faces are removed based on their normals: faces are discarded if their normal vector forms an angle of less than a threshold with the wall's normal, depicted as \textit{Filtered by Normal} in \cref{fig:filteringScheme}. These thresholds used for mesh filtering are empirically determined and are reported in \cref{subsec:setup}.
The unfiltered region of $M_{\text{original}}$, is specified as \textit{Intact Area} in \cref{fig:filteringScheme}. The resulting filtered mesh, $M_{\text{filtered}}$ is illustrated in \cref{fig:filteredMesh}. \\
Finally, we perform the remeshing process, shown in \cref{fig:remeshingScheme}. Initially, every vertical wall is composed of two triangular faces. These walls are then remeshed by inserting grid points, using Constrained Delaunay Triangulation (CDT)~\cite{CDT}. CDT edges are inserted to encircle vertical planes. In the case of closed rooms, the floor and ceiling are formed by CDT by inserting constraining edges around the closed loop. Moreover, evenly distributed grid points and along-edge points are added within the loop. For open rooms, the convex hull shown in \cref{fig:mbbAndCDT} is used to form the floor and ceiling, although it may not accurately represent the room's contour.
The remeshed vertical and horizontal planes are integrated with $M_{\text{filtered}}$ to form $M_{\text{combined}}$, shown in \cref{fig:combinedMesh}. During this process, semantic information is precisely documented, linking each mesh segment to its respective semantic component. $M_{\text{combined}}$ is the input to the texturing phase detailed in \cref{SystemDesign:Texturing}. More visualizations of the Mesh Processing step are shown in the supplementary (Section A.1).

\vspace{-4pt}
\subsection{Texture Images Acquisitions }
\label{SystemDesign:TextureImageAcquAndTexturing}
As outlined in \cref{SystemDesign:problemStatement}, texturing algorithms \cite{SEAMLESS, MVSTexturing} encounter issues such as blurry textures (from using $C_{\text{original}}$) and visible texture seams. To address these challenges, we introduce an AR-Guided texture image capturing approach specifically designed for texturing vertical walls. This method offers three key advantages. First, it allows users to capture images in a more stable manner, producing sharper images. Second, it seeks to minimize the number of images required per plane, as texturing with a single image reduces the misalignment issues associated with texturing using multiple images. Lastly, this strategy significantly speeds up the texturing process due to fewer images being used. When capturing large planes, multiple images may be needed due to limitations in camera field of view (FoV) or obstructions. To overcome this, we introduce the \textbf{Divide and Conquer} algorithm, which segments a large plane into smaller subplanes that can each be covered by a single image, whilst minimizing the overall number of images required.

\subsubsection{Large Plane Dividing (Divide)}
\begin{figure}[tb]
 \centering
 \includegraphics[width=\columnwidth]{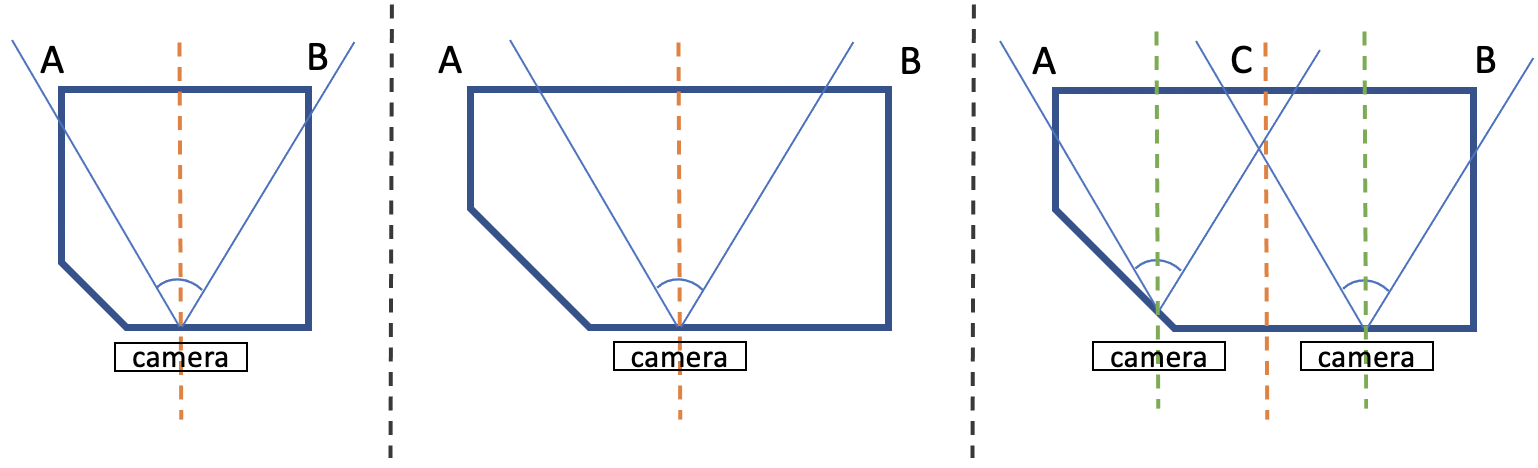}
 \vspace{-20pt}
 \caption{\textbf{Splitting scheme for large planes}: In the left image, a small plane is adequately covered by a single image. The middle image illustrates a large plane that extends beyond the scope of one image. In the right image, the large plane (AB) is divided into sub-planes (AC and BC), each of which can be completely covered by a single image.}
 \vspace{-15pt}
 \label{fig:method:splitting}
\end{figure}
We develop a plane splitting algorithm that assesses the feasibility of capturing an entire plane in one shot as the user navigates the space (see \cref{fig:method:splitting}). Using the phone's FoV in portrait mode, determined by $K_{\text{average}}$, we calculate the optimal distance $d^*$ from which the camera should capture the plane of interest. From this, the optimal camera position, $\vec{X^*}$ is calculated as follows: $\vec{X^*} = \vec{c_{p}} + d^* \times \vec{n}_{p}$, where $\vec{c_{p}}$ and $\vec{n}_{p}$ are the plane of interest's center position and normal vector, respectively. With a standard ceiling height of $2.44\ \text{m}$~\cite{ceilingHeight}, we propose that aligning the optimal camera height (in portrait mode capture) with the midpoint of the wall height (i.e. around $1.22\ \text{m}$) is ergonomically suitable for users. Then, we verify whether $\vec{X^*}$ remains within the room layout. If $\vec{X^*}$ is deemed too far or impractical, we recursively subdivide the plane into smaller subplanes, until the optimal conditions for capture are met. This recursion is guided by two critical criteria: the distance from the wall, crucial for capturing detailed and realistic images, and polygon inlier status, which checks if the camera position is within the room's polygonal boundary. All in all, our \textbf{divide} algorithm allows for capturing plane sections at closer and more effective distances. Refer to the supplementary material for more details (Section B, Algorithm 3).

\begin{figure}[ht]
    \centering
    \begin{subfigure}[b]{0.39\columnwidth}
        \centering
        \includegraphics[width=0.8\textwidth]{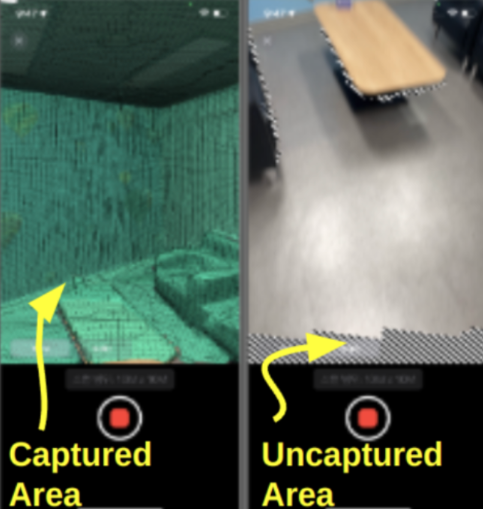}
        \caption{\textbf{UI for the Capturing Step}. The left view shows captured area and the right view shows uncaptured area.}
        \label{fig:method:captureView}
    \end{subfigure}
    \hfill
    \begin{subfigure}[b]{0.58\columnwidth}
        \centering
        \includegraphics[width=1.0\textwidth]{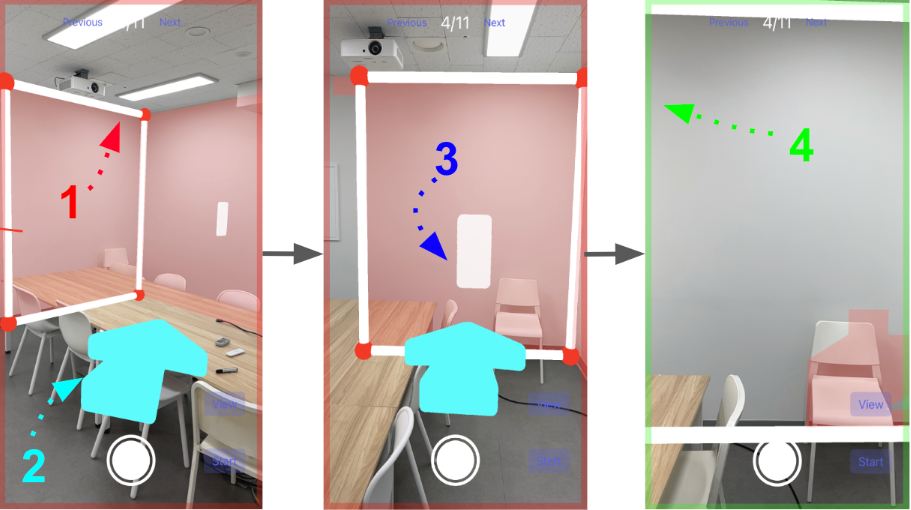}
        \caption{\textbf{UI for Divide and Conquer AR-Guided image capturing step}: guides are the target plane (1), optimal camera position indicators (2,3), and a green boundary (4) for proximity alerts.}
        \label{fig:method:UIForDivisionAndConquer}
    \end{subfigure}
    \caption{UI for the \textbf{Capturing} and \textbf{Conquer step.}}
    \vspace{-15pt}
    \label{fig:UIGroup}
\end{figure}

\subsubsection{Conquer}
The \textbf{conquer} step aims to guide the user to the optimal $\vec{X^*}$ to capture images for texturing. For this, we introduce an AR guidance system that directs the user to $\vec{X^*}$. As shown in \cref{fig:method:UIForDivisionAndConquer}, the target subplane, marked as (1), is emphasized by a red box with white edges. The user is directed towards the optimal capture position by a cyan arrow, marked as (2) in \cref{fig:method:UIForDivisionAndConquer}. $\vec{X^*}$, marked as (3) in \cref{fig:method:UIForDivisionAndConquer}, is visualized in AR as a white rectangular 3D phone model. The device interface displays a green boundary (marked as (4) in \cref{fig:method:UIForDivisionAndConquer}) and provides haptic feedback to signal proximity to the target position.
These guides ensure precise user alignment with the target $\vec{X^*}$. Upon capturing a target subplane, it reverts to its natural color, confirming success, whilst uncaptured subplanes are left in red.
Images obtained through this process are designated as $C_{\text{recaptured}}$, marking the completion of the first phase of our pipeline.

\subsection{Texturing}
\label{SystemDesign:Texturing}
Starting with the second phase of our pipeline, the \textbf{Texturing} process aims to give realism to the combined mesh $M_{\text{combined}}$ from \cref{SystemDesign:MeshProcessing}. We utilize SOTA assignment-based texturing methods such as SeamlessTex~\cite{SEAMLESS} and MVSTex~\cite{MVSTexturing}. However, instead of using $C_{\text{original}}$ from the Capturing stage (\cref{SystemDesign:Capturing}), our method employs $C_{\text{recaptured}}$ from \cref{SystemDesign:TextureImageAcquAndTexturing}, tailored for texturing. For further details on the texturing algorithms employed, please refer to the appendix (Section A.4) and relevant literature~\cite{SEAMLESS,MVSTexturing,advertinit}.
The resulting textured mesh, $M_{\text{textured}}$, includes both the textured room layout mesh (textured version of the remeshed room layout from \cref{SystemDesign:MeshProcessing}) and the textured object mesh (textured version of $M_{\text{filtered}}$ from \cref{SystemDesign:MeshProcessing}). Utilizing semantic information retrieved during the Mesh Processing step, these components are systematically divided into distinct textured meshes for further processing.

\vspace{-3pt}
\subsection{Plane2Image}
\label{SystemDesign:Plane2Image}
\begin{figure}[tb]
 \centering
 \includegraphics[width=1.0\columnwidth,height=0.35\columnwidth]{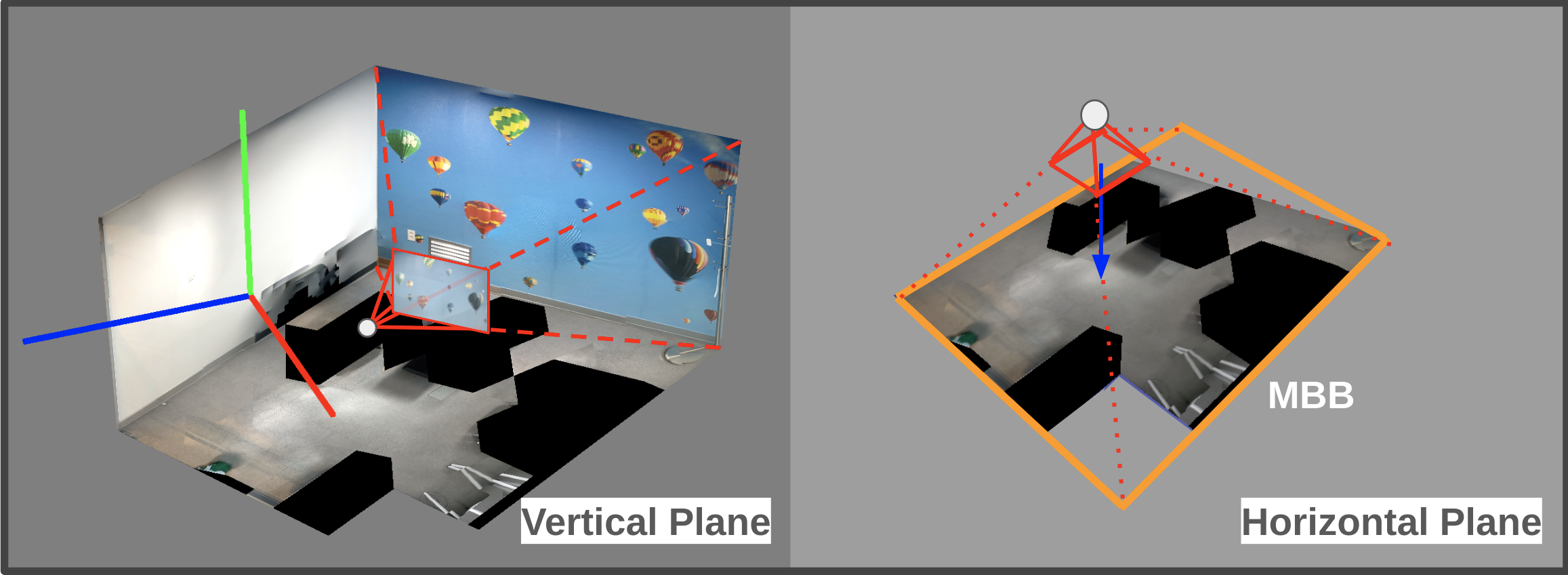}
 \caption{Plane2Image rendering operation for vertical and horizontal planes.}
 \label{fig:method:plane2image}
\end{figure}

The aim of \textbf{Plane2Image} is to render 2D images from 3D textured planes as illustrated in \cref{fig:method:plane2image}. To do so, we use the semantically split textured meshes from \cref{SystemDesign:Texturing} to ensure rendering without interference, and leverage the Metal~\cite{Metal} (for iOS) and OpenGL~\cite{OpenGL} (for Linux) frameworks. For vertical planes, we extract information of each plane such as the normal vector, center, width, and height from the database. A virtual camera is then positioned and oriented based on these parameters to precisely render the plane to a 2D image.
For horizontal surfaces, we instead use the MBB parameters from \cref{SystemDesign:MeshProcessing}. For more details on this rendering process, please consult the supplementary material (Section B, Algorithm 1 and Algorithm 2). This transition from 3D to 2D enables the application of various image processing techniques, such as image inpainting~\cite{ZIT++, ZIT, MAT, LamaInpainting} and generative AI methods~\cite{Photoshop, DALLE, LDM, Zhuang2023ATI}, as further outlined in \cref{SystemDesign:PostTexturing}.
It is important to note that the remeshed vertical wall meshes from \cref{SystemDesign:MeshProcessing} are reverted back to their original two-face configuration, reducing the mesh's number of vertices and faces. This simplification aids in texture coordinate mapping. For vertical planes, vertices are assigned new texture coordinates based on only the four plane corners. Rendered images are then applied using these texture coordinates for the simplified faces. Horizontal surfaces undergo a similar face and vertex reduction, where the texture coordinates are mapped based on the MBB parameters.

\begin{figure}[tb]
 \centering
 \includegraphics[width=1.0\columnwidth]{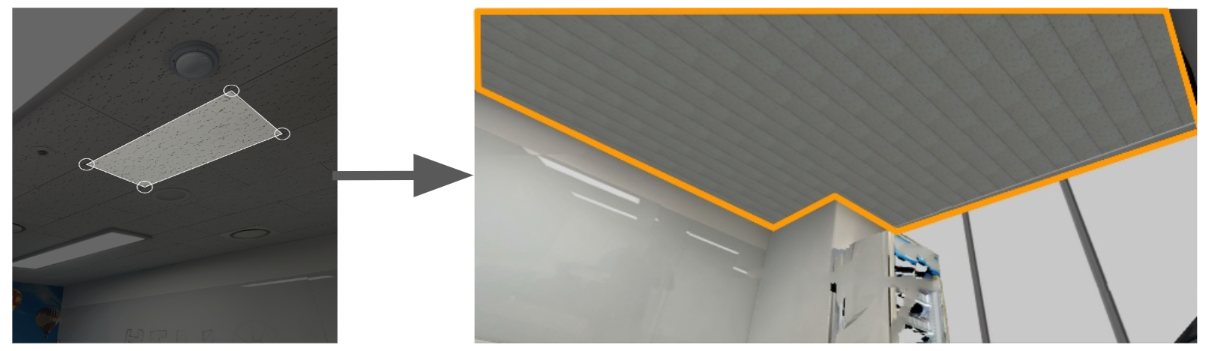}
 \caption{UI for capturing a ceiling sample (left), and a ceiling texture created using the captured sample (right).}
 \label{fig:postTexturingSample}
\end{figure}

\begin{table*}[ht]
  \caption{Results of local and global plane-to-plane texture quality assessments (standard deviations in parentheses).}
  \label{tab:combined_experiments}
  \centering
  \resizebox{\textwidth}{!}{%
    \begin{tabular}{c|cccc|cccc}
      \toprule
      & \multicolumn{4}{c|}{ Local Texture Quality Eval.} & \multicolumn{4}{c}{Global Plane-to-Plane Texture Quality Eval.} \\
      \midrule
      Method & PSNR[dB]$\uparrow$ & SSIM[-]$\uparrow$ & ST-LPIPS[-]$\downarrow$ & Blurriness[-]$\downarrow$ & PSNR[dB]$\uparrow$ & SSIM[-]$\uparrow$ & ST-LPIPS[-]$\downarrow$ & Blurriness[-]$\downarrow$ \\
      \midrule
      ColorMapOpt~\cite{COLORMAPOPTIM} & 20.19 ($\pm$1.48) & 0.84 ($\pm$0.02) & 0.17 ($\pm$0.02) & 0.25 ($\pm$0.04) & 16.09 ($\pm$1.64) & 0.77 ($\pm$0.06) & 0.25 ($\pm$0.06) & 0.44 ($\pm$0.05) \\
      PlaneOpt~\cite{PlaneOpt} & 20.13 ($\pm$0.99) & 0.84 ($\pm$0.02) & 0.17 ($\pm$0.02) & 0.22 ($\pm$0.04) & 16.49 ($\pm$1.72) & 0.77 ($\pm$0.06) & 0.25 ($\pm$0.06) & 0.41 ($\pm$0.04) \\
      MVSTex~\cite{MVSTexturing} & 20.87 ($\pm$1.79) & \textbf{0.86} ($\pm$0.03) & 0.10 ($\pm$0.01) & 0.22 ($\pm$0.03) & 17.09 ($\pm$2.12) & 0.80 ($\pm$0.06) & 0.19 ($\pm$0.04) & 0.33 ($\pm$0.02) \\
      SeamlessTex~\cite{SEAMLESS} & 21.32 ($\pm$1.61) & \textbf{0.87} ($\pm$0.03) & 0.09 ($\pm$0.02) & 0.23 ($\pm$0.03) & 17.61 ($\pm$2.18) & 0.81 ($\pm$0.06) & 0.18 ($\pm$0.04) & 0.34 ($\pm$0.03) \\
      Our-MVSTex & \textbf{21.35} ($\pm$1.40) & 0.86 ($\pm$0.03) & \textbf{0.09} ($\pm$0.02) & \textbf{0.20} ($\pm$0.02) & \textbf{20.35} ($\pm$1.94) & \textbf{0.91} ($\pm$0.02) & \textbf{0.11} ($\pm$0.02) & \textbf{0.29} ($\pm$0.02) \\
      Our-SeamlessTex & \textbf{21.33} ($\pm$1.40) & 0.86 ($\pm$0.03) & \textbf{0.09} ($\pm$0.02) & \textbf{0.20} ($\pm$0.02) & \textbf{20.80} ($\pm$2.27) & \textbf{0.92} ($\pm$0.02) & \textbf{0.10} ($\pm$0.02) & \textbf{0.29} ($\pm$0.02) \\
      \bottomrule
    \end{tabular}%
  }
  \vspace{-7pt}
\end{table*}

\subsection{Post Texturing}
\label{SystemDesign:PostTexturing}
In \cref{SystemDesign:Plane2Image}, we established one-to-one correspondences between 3D planar structures and 2D images, simplifying the process of improving textures to merely switching images or altering them. In this section, we introduce various texturing enhancement methods that are performed by directly modifying information in the 2D image domain. Mainly, we propose two post texturing techniques, the \textbf{ImageMode} and the \textbf{SampleMode}, applicable to both vertical and horizontal surfaces.\\
The ImageMode allows users to replace the defective rendered 2D images from the Plane2Image phase with any image of their choice, offering high versatility and customizability. Specifically, image inpainting models~\cite{ZIT++, ZIT, MAT, LamaInpainting} successfully address both the environmental and technological challenges by filling in untextured areas (from occlusions), correcting texture artifacts, and removing undesired contents. This significantly enhances the overall quality and accuracy of the texturing outputs, as shown in \cref{fig:qualitativeBigFigure} for the ZIT inpainting model~\cite{ZIT} used by RoomRecon. ZIT inpainting is chosen as it provides the best trade-off between quality and inference speed relative to other methods~\cite{MAT, LDM, LamaInpainting}. For a detailed rationale behind this choice, please refer to the supplementary material (Section A.5). On top of inpainting, stable diffusion-based methods~\cite{LDM, Zhuang2023ATI} allow for the addition of creative content to original textures, allowing customization of room decorations according to user preferences. For additional visualizations, please refer to the supplementary material (Section A.5).
The SampleMode targets the challenges of achieving high-quality textures for floors and ceilings. Users have the option to select a sample from a database or capture one from real life. Therefore, we provide a UI for capturing samples from these surfaces as shown in \cref{fig:postTexturingSample}. Users can customize their floor and ceiling textures by adjusting parameters such as sample width, sample height, sample angle and sample offset, allowing for the creation of textures that closely resemble the real world. An example of a ceiling texture, created using our SampleMode, is shown in \cref{fig:postTexturingSample}. For further visualization of textures created using SampleMode, refer to the supplementary material (Section A.2). This marks the end of the RoomRecon pipeline.

\section{Experiments}
\label{Experiments}

In this section, we assess the performance of various texturing methods within our RoomRecon pipeline, with a focus on both local and global texture quality, and time complexity.
For a complete and comprehensive result of our experiments, please consult the supplementary materials (Section C).

\begin{table}[]
\vspace{-5pt}
\caption{Blurriness scores of original, sampled and recaptured images (left). Processing time of different texturing methods (right).}
\vspace{-5pt}
\centering
\resizebox{\columnwidth}{!}{%
\begin{tabular}{lccc|ll}
\toprule
 & Orig. & Sampled & Recapt. & Methods & Time[s] \\ 
\midrule
Num. of  & 599.92 & 258.85 & 11.38 & ColorMapOpt & 78.50($\pm$ 35.48) \\
Frames[-] & ($\pm$ 262.90) & ($\pm$ 106.98) & ($\pm$ 3.82) & PlaneOpt & 273.79($\pm$ 179.96) \\
\cline{1-4}
Blur [-] & 0.41 & 0.38 & 0.25 & MVSTex & 8.48($\pm$ 2.03) \\
 & ($\pm$ 0.06) & ($\pm$ 0.05) & ($\pm$ 0.02) & SeamlessTex & 32.62($\pm$ 18.93) \\
\cline{1-4}  
 &  &  &  & Our-MVSTex & 3.23($\pm$ 1.05) \\
 &  &  &  & Our-SeamlessTex & 4.44($\pm$ 1.29) \\
\bottomrule
\end{tabular}%
}
\label{tab:blurAnalysisAndTimeOfTexturing}
\vspace{-15pt}
\end{table}

\subsection{Experiment Setup}
\label{subsec:setup}
In our paper, we conduct experiments within diverse indoor spaces, including seven rooms of office environments and six rooms of household environments. The data acquisition, involving RGB-D images and camera parameters, is performed using the iPhone 12 Pro, integrated with the ARKit framework~\cite{ARKIT}. Room layouts are obtained through Apple Inc.'s API~\cite{AppleRoomPlan}. Detailed specifications for each room are available in the supplementary material (Section C).
We compare the texture capabilities of six texturing methods, reimplemented within our RoomRecon texturing pipeline: ColorMapOpt~\cite{COLORMAPOPTIM}, PlaneOpt~\cite{PlaneOpt}, MVSTex~\cite{MVSTexturing}, SeamlessTex~\cite{SEAMLESS}, and our adaptations, Our-MVSTex and Our-SeamlessTex.  We refer to our adaptations as \textit{Ours} from now onwards. SeamlessTex~\cite{SEAMLESS} originally includes camera pose optimization and texture synthesis processes, which we omit in our adaptation due to their high computational demands. 
\textit{Ours} utilizes newly captured images $C_{\text{recaptured}}$ for texturing, whereas standard algorithms such as MVSTex, SeamlessTex, ColorMapOpt, and PlaneOpt use sampled images $C_{\text{sampled}}$, selected from $C_{\text{original}}$ every four timesteps based on sharpness. Typically, the other three images are discarded unless their poses significantly differ from the selected image. All other aspects of the RoomRecon pipeline remain identical across all experiments. \\
In the Capturing phase, we set the voxel size to $20 \, \text{mm}$ and utilize $192 \times 256$ RGB-D images, leveraging camera parameters provided by ARKit \cite{ARKIT}. Real-time mesh generation is facilitated through the Metal framework \cite{Metal}. The initial mesh ($M_{\text{original}}$) undergoes a $50\%$ reduction in face count using Garland's simplification algorithm \cite{quadricSimpl}.
During the mesh filtering scheme in \cref{SystemDesign:MeshProcessing}, faces situated within $0-5 \, \text{cm}$ from a plane are excluded via the \textit{filtered by distance} approach, while those within $5-10 \, \text{cm}$ are filtered using the \textit{filtered by normal} strategy, with a threshold angle of $10\ \text{degrees}$.
In the remeshing stage of \cref{SystemDesign:MeshProcessing}, we specify a grid resolution of five points per meter for vertical planes and a uniform grid of $15 \times 15$ points throughout the MBB for horizontal planes. 
For the Plane2Image stage, rendering parameters are configured such that each meter of the plane corresponds to $500$ pixels in the resulting image. As an illustration, a plane measuring $2 \, \text{m}$ in width and $1 \, \text{m}$ in height is rendered as an image with dimensions $1000 \times 500$ pixels. \\
Local Texture Quality Evaluation (\cref{res: local eval}) and Global Plane-to-Plane Evaluation (\cref{res: global eval}) experiments are conducted prior to the application of inpainting or any Post Texturing processes.
For both evaluations, the focus is exclusively on textures of vertical walls, excluding floors, ceilings, and object textures. This is because floors and ceilings are regenerated using the SampleMode in the Post Texturing step (Sec. 3.8), and our focus is on texturing structural elements, not objects.

\subsection{Local Texture Quality Evaluation}
\label{res: local eval}
\textbf{Methodology:} The objective is to assess local texture quality of walls, with the following evaluation procedure. Ground truth (GT) images and corresponding camera parameters are captured using the RoomRecon framework, focusing on semantically important regions within the room that have diverse colors, lines and edges (e.g.: wall paintings, doors and other highly detailed areas). We avoid uniform regions such as plain white walls, where texturing errors are hard to spot. For samples of these captured ground truth images, refer to the supplementary material (Sec. A.6). The camera intrinsics and extrinsics are tracked by ARKit~\cite{ARKIT}. Each texturing method is applied to the room mesh, and the resulting textured meshes are rendered at the corresponding GT poses (i.e. poses at which the GT images were captured) for comparison in the 2D image space. This evaluation is referred to as \textit{local texture quality assessment} because it examines only the specific, limited areas of the scene captured by the purposely curated GT. Consequently, the evaluation focuses on partial areas, whether it be portions of a single plane or encompassing multiple planes. 
Masked GT images and masked rendered images are compared for four metrics, namely PSNR~\cite{PSNR}, SSIM~\cite{SSIM}, Shift Tolerant Learned Perceptual Image Patch Similarity (ST-LPIPS)~\cite{stlpips}, and Blurriness~\cite{Blurriness}. These metrics are selected, as done by prior work~\cite{SEAMLESS, advertinit}, to ascertain the similarity of textured rendered outputs to GT images (PSNR, SSIM, and ST-LPIPS), and to evaluate the sharpness and vividness of the textured outputs using the Blurriness measure, which is GT-independent. Specifically, PSNR compares pixel values and ranges from $0-60\ \text{dB}$, whilst SSIM compares image properties (e.g.: luminance, contrast and structure) on a scale from $0-1$.\\
\textbf{Discussion:}
The left section of \cref{tab:combined_experiments} highlights the performance of six texturing methods on local texture quality. Our methods outperform all previous methods across most metrics except in SSIM and ST-LPIPS, where the scores are comparable to those of the baseline methods: SeamlessTex~\cite{SEAMLESS} and MVSTex~\cite{MVSTexturing}.
This similarity in SSIM and in ST-LPIPS can be attributed to the fact that SeamlessTex and MVSTex try to use the same image to texture groups of neighboring mesh faces. Hence, in the context of our local texture quality assessment, it is highly likely that all meshes are textured with the same image, leading to few seams and structural discrepancies. In this setting, our methods have small margins for improvement, as we also try to minimize the number of images used for texturing. \\
On the other hand, our techniques demonstrate a significant advantage in PSNR and Blurriness. As shown in \cref{tab:blurAnalysisAndTimeOfTexturing}, the original set of color images $C_{\text{original}}$ exhibits a high Blurriness score of 0.41, and even the sharpest sampled $C_{\text{sampled}}$, used by the prior methods, only improve the Blurriness to 0.38. In contrast, the $C_{\text{recaptured}}$ through our Divide-And-Conquer scheme are $39\%$ and $34\%$ sharper than the original and sampled images, respectively. This leads to a $9\%$ and $13\%$ sharper texture output compared to MVSTex and SeamlessTex respectively, as our \(C_{\text{recaptured}}\) images are utilized for the texturing process.
Additionally, our superior performance in PSNR demonstrates that our methods more closely resemble the GT images in terms of pixel values. 
Finally, we notice that our methods perform similarly across all metrics, irrespective of the underlying texturing method. This shows that using few, high-quality input images to the texturing module is of upmost importance, reducing the need for complicated texturing algorithm.

\vspace{-5pt}
\subsection{Global Plane-to-Plane Texture Evaluation}
\label{res: global eval}
\begin{figure*}[ht]
  \centering
\includegraphics[width=0.8\linewidth, height=0.90\linewidth]{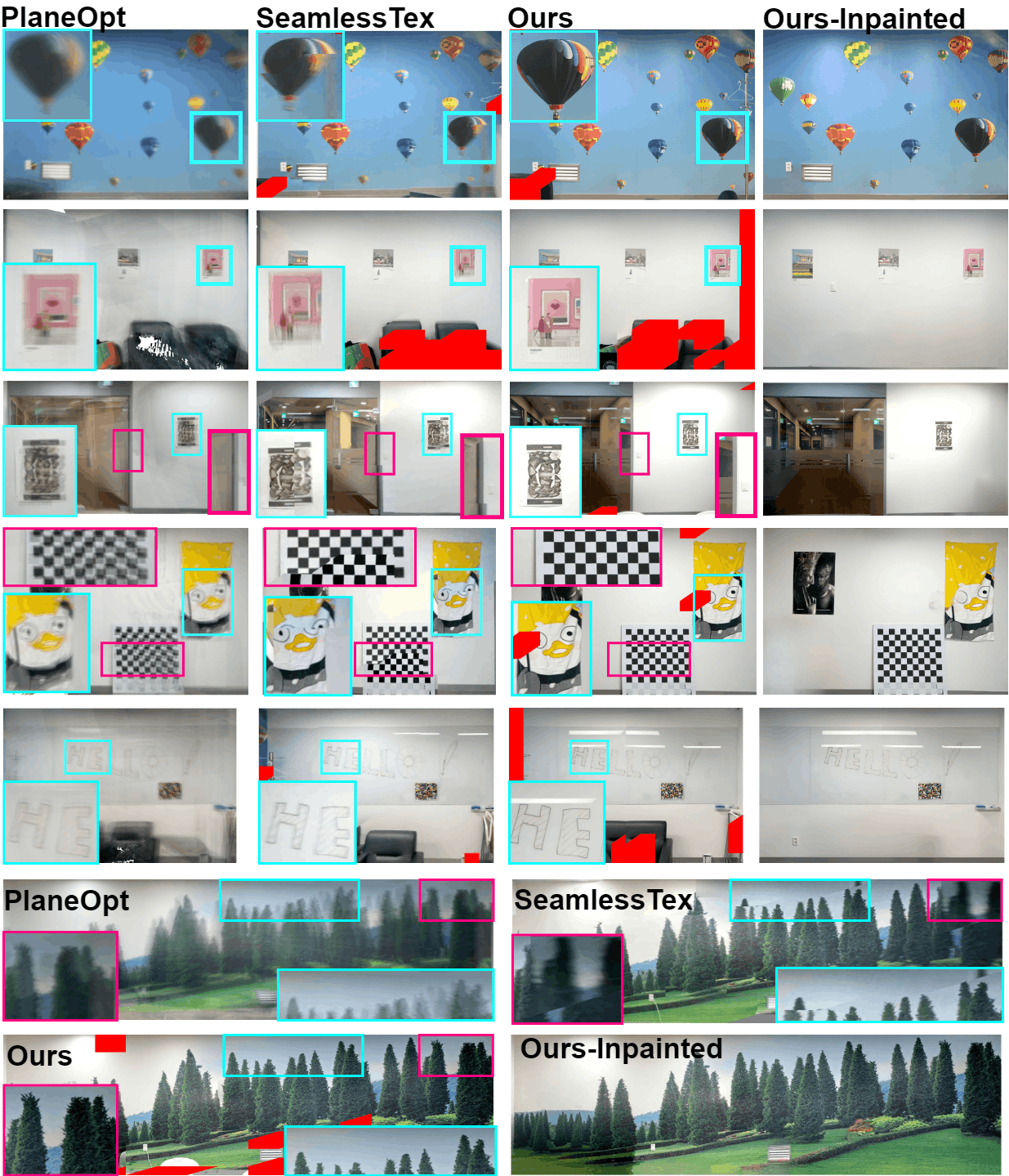}
  \caption{Qualitative results of textured planes using PlaneOpt~\cite{PlaneOpt}, SeamlessTex~\cite{SEAMLESS}, Our-SeamlessTex, and Our-SeamlessTex after inpainting operation. For more visualizations of results using Ours-Inpainted, refer to the supplementary material (Section A.7).
  }
\label{fig:qualitativeBigFigure}
\end{figure*}

\begin{figure*}[ht]
  \centering
  \includegraphics[width=0.8\linewidth]{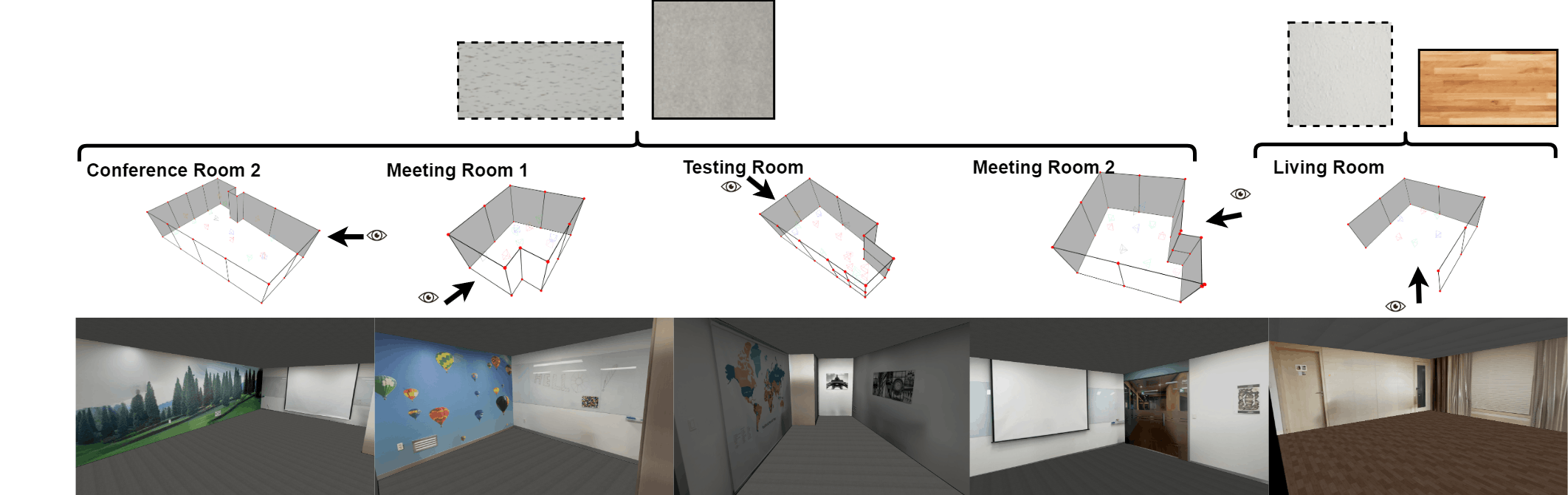}
  \caption{Textured room layouts after completing our RoomRecon pipeline. Samples with dashed lines show ceiling samples used for creating ceiling textures and samples in solid lines show floor samples for floor textures. For additional viewpoints of the textured layouts, refer to the supplementary material (Section A.7).
  }
\label{fig:finalResult}
\end{figure*}

\textbf{Methodology:} This study compares entire plane textures against high-quality GT images. GT images of full walls are captured with mobile phones, employing techniques such as 0.6x zoom and panorama shots. These images undergo a homography transform, where the four vertices of the walls are manually clicked to create GT planar images for the respective walls. Then, the images are refined using Adobe Photoshop~\cite{Photoshop} to establish high-quality benchmarks for comparison, including the removal of furniture from images. During the evaluation, the rendered images of textured planes from the Plane2Image stage are compared to these GT images. The evaluation utilizes the same four metrics employed in \cref{res: local eval}.\\
\textbf{Discussion:}
The right section of \cref{tab:combined_experiments} presents the performance of six texturing methods within a global plane-to-plane evaluation framework. \textit{Ours} demonstrate significant improvements across all metrics compared to prior work. For example, for Our-SeamlessTex, PSNR improves by $18.1\%$, SSIM by $13.6\%$, ST-LPIPS by $44.4\%$, and Blurriness by $14.71\%$, compared to its baseline SeamlessTex. The Blurriness reductions and PSNR improvements align with findings from the local evaluation in \cref{res: local eval}, highlighting the effect of using sharper images. Moreover, unlike in the local results, our methods greatly surpass their baselines in SSIM and ST-LPIPS. This demonstrates that using fewer, carefully captured images through our divide-and-conquer scheme, leads to rendered images that more closely replicate the GT image properties in terms of structure and color. Moreover, as each image captures a significant portion of a wall, it ensures that all structures and details of that wall area are intact and free of texture seams. Again, the similar scores between our methods suggests that using quality input data for texturing removes the need for highly complex texturing algorithms. For more qualitative comparisons of textured results using PlaneOpt, SeamlessTex, and Our-SeamlessTex, along with ZIT-inpainted~\cite{ZIT} version of Our-SeamlessTex, refer to \cref{fig:qualitativeBigFigure}. This figure illustrates the visual enhancements and the detailed realism achieved through our texturing methods, providing a clearer understanding of the quantitative results discussed above. Full rooms created following our RoomRecon pipeline with Our-SeamlessTex are visualized in \cref{fig:finalResult}.

\vspace{-3pt}
\subsection{Time Analysis of Texturing Methods and RoomRecon Pipeline}
We examine the texturing times of various methods on a Linux machine with AMD Ryzen 9 5900x 12-core × 24 processor and report our findings in the right part of \cref{tab:blurAnalysisAndTimeOfTexturing}. Our approach considerably accelerates the texturing process, achieving times of just $3.23\ \text{s}$ and $4.44\ \text{s}$ for our-MVSTex and our-SeamlessTex respectively. This is a notable increase in speed compared to their baselines that run in $8.48\ \text{s}$ and $32.62\ \text{s}$ respectively. Our method achieves increased efficiency primarily through the implementation of our AR-Guided image capturing approach. This method minimizes the number of captures required to cover a room, resulting in an average of 11.38 images per room, a significant reduction from the typical 258.85 images needed (\(C_{\text{sampled}}\)), as stated in the left side of \cref{tab:blurAnalysisAndTimeOfTexturing}. Using fewer images for texturing simplifies the selection of optimal images for each face, leading to shorter texturing time.
Additionally, we provide a breakdown of the time each step takes in our RoomRecon pipeline using the standard MVSTex~\cite{MVSTexturing} and Our-MVSTex on an iPhone 12 Max, as outlined in \cref{tab:timeOnIOS}. The pipeline steps common to both texturing methods include the room layout parsing, Mesh Processing (\cref{SystemDesign:MeshProcessing}), Plane2Image (\cref{SystemDesign:Plane2Image}), and the display of textured meshes (\(M_{\text{textured}}\)) on the phone's screen. These combined common processes take an average of $4.8\ \text{s}$ to compute. As RoomRecon's texturing module differs from prior works, a time analysis of MVSTex \cite{MVSTexturing} and our-MVSTex is also outlined in \cref{tab:timeOnIOS}. In the implementation of MVSTex~\cite{MVSTexturing}, the texturing step involves calculating the Blurriness score for each image in  \(C_{\text{original}}\) to sample only the sharpest images for texturing (\(C_{\text{sampled}}\)). This step and the actual MVSTex take an average of $4.85\ \text{s}$ and $18.71\ \text{s}$ respectively. With our new Divide and Conquer scheme, we eliminate the need to calculate blurriness, as we recapture few sharp images for texturing. The Divide stage is practically instantaneous ($0.23\ \text{ms}$), and the significantly fewer images used for our texturing (from $258.85$ to $11.38$) lead to a total texturing time of just $4.72\ \text{s}$. Thus, RoomRecon with our-MVSTex is processed in $9.48\ \text{s}$, whereas RoomRecon with MVSTex runs three times slower.  
Our simplified approach reduces waiting time for users as their models are processed more rapidly. However, it's important to acknowledge that our method incorporates a second capturing step, not factored into the time analysis provided, which solely focused on algorithm processing time. This additional image acquisition stage introduces a time overhead, dependent upon the user's familiarity with the RoomRecon application. Therefore, the overall time required for an experienced user to generate a virtual textured room remains comparable irrespective of the method used, yet our approach places less strain on the phone.

\subsection{User Evaluation}

\noindent
We conduct a user study to assess how different people judge the quality of different texturing methods. After filling in a consent form, a survey is given to each participant, with the task of selecting the \textit{best} rendered image amongst a set of rendered images textured by different methods. We explain to participants to judge based on uniform and consistent illumination, texture seams and clarity. Specifically, we randomly select $20$ plane images, each textured with these four texturing methods: MVSTex~\cite{MVSTexturing}, SeamlessTex~\cite{SEAMLESS}, our-MVSTex, and our-SeamlessTex, such that a total of $80$ samples are provided. A group of four images for each scene are randomly displayed. ZIT inpainting~\cite{ZIT} is utilized to fill untextured areas in all rendered images. For each scene, we show the four differently textured images side by side and ask the participants to record their preferences. 

\begin{table}[htbp]
\centering
\caption{Detailed process and texturing time analysis on iPhone for MVSTex~\cite{MVSTexturing} and Our-MVSTex.}
\vspace{-5pt}
\label{tab:timeOnIOS}
\resizebox{\columnwidth}{!}{%
\begin{tabular}{lrr}
\toprule
\textbf{Common Process (a)} &  \multicolumn{2}{c}{}  \\
\midrule
Layout Parsing [s] & \multicolumn{2}{c}{$(3.70 \pm 1.30)\mathrm{e}{-3}$} \\
Mesh Proc. [s] & \multicolumn{2}{c}{$0.23 \pm 0.17$} \\ 
Plane2Img [s] & \multicolumn{2}{c}{$3.93 \pm 1.46$} \\ 
Display Results [s] & \multicolumn{2}{c}{$0.60 \pm 0.35$} \\ 
\midrule
\textbf{Texturing Process (b)} & \textbf{Original MVSTex.} & \textbf{Divide and Conquer (Ours)} \\
\midrule
Frames [-] & $599.92 \pm 262.90$ & $11.38 \pm 3.82$ \\
Blur Calc. [s] & $4.85 \pm 2.24$ & -- \\
Divide Calc. [s] & -- & $(0.23 \pm 0.04)\mathrm{e}{-3}$ \\
Texturing Time [s] & $18.71 \pm 9.25$ & $4.72 \pm 2.06$ \\
\midrule
RoomRecon Total (a + b) [s] & $28.33 \pm 12.77$ & $9.48 \pm 3.69$ \\
\bottomrule
\end{tabular}
}
\vspace{-5pt}
\end{table}

\begin{table}[htbp]
\caption{User study votes for the best texture solution.}
\vspace{-5pt}
\begin{adjustbox}{width=0.9\columnwidth,center}
  \centering
  \begin{tabular}{l|cccc}
    \toprule
    Method & MVSTex & SeamlessTex & our-MVSTex & our-SeamlessTex \\
    \midrule
    Number of votes & 45(7.5\%) & 35(6.8\%) & 262(43.6\%) & 258(42.1\%)  \\
    \bottomrule
  \end{tabular}
\end{adjustbox}
\label{tab:user_study}
\vspace{-7pt}
\end{table}

\noindent
A group of 30 people with different backgrounds are asked to complete the survey. The age and background distribution are as follows: 14 software engineers, 6 designers, and 10 computer science students. There are 18 males (average age $33.3\pm5.0$) and 12 females (average age $30.2\pm4.1$), with an overall average age of $32.1\pm4.8$ years. In total, 600 selections are recorded (1 per person per scene), and the voting results for each texturing method are summarized in \cref{tab:user_study}. Most participants chose our-MVSTex and our-SeamlessTex as the best methods, with voting rates of 43.6\% and 42.1\% respectively. This shows the qualitative superiority of our textured meshes, emphasizing the positive effect of our proposed techniques on texture quality, clarity and seam reduction.

\label{Result:Result}

\vspace{-3pt}
\section{Limitations and Future Work}
\label{FutureWork}

Whilst RoomRecon successfully demonstrates better texturing quality compared to previous methods, it presents some limitations. Firstly, our divide-and-conquer capturing method is restricted in narrow corridors as it cannot find suitable capture positions inside the room. Hence, it is essential to further refine the algorithm to handle such edge cases. Next, the current prototype does not optimize for user interaction time, which is important for the overall user experience. Lastly, the absence of a thorough user evaluation limits the understanding of the proposed UI’s usability across different user demographics. The last two limitations come from the fact that this paper focuses on presenting a pipeline to improve texturing quality. We suggest a UI configuration to prove that involving users can improve texturing. However, our UI represents one potential approach, and we believe that the design and evaluation of an effective UI considering speed and ease of use can be the target of future work. Aside from tackling the above points, future work can focus on expanding RoomRecon's capabilities to multi-room settings, allowing for the capture of entire floors (see supplementary material, Section A.3). Moreover, integrating practical requirements from experts in target fields such as real estate and interior design would improve RoomRecon's performance and 3D modeling capabilities in these target applications.

\section{Conclusion}
\label{Conclusion}

In this study, we introduced RoomRecon, a novel mobile application designed for the precise and realistic texturing of indoor spaces. By leveraging the capabilities of mobile devices and focusing on the static elements of rooms, RoomRecon allows users to generate detailed 3D models with minimal need for updates. This application tackles limitations of current 3D reconstruction technologies by offering an AR-guided UI that supports the capturing experience. Through a dedicated two-phase pipeline, RoomRecon addresses common texturing issues such as blurring and seams, outperforming traditional methods both quantitatively and qualitatively.

\clearpage
\bibliographystyle{abbrv-doi}
\bibliography{template}

\clearpage
\begin{center}
{\LARGE\bfseries Overview of Supplementary Material}
\end{center}

\begin{enumerate}
    \item Section A: We provide further visualizations of the RoomRecon processes and results.
    \item Section B: We provide pseudocodes that explain the related algorithms in the main paper.
    \item Section C: We present extensive tables of experiments showing quantitative statistics on the test datasets.
\end{enumerate}

\section{Further Visualizations of the\\
        RoomRecon Processes and Results}
\label{Supp: moreVis}

\subsection{Detailed visualizations of the Mesh Processing step}
\begin{figure}[h]
\centering
\begin{subfigure}{0.5\columnwidth}
  \includegraphics[width=\linewidth]{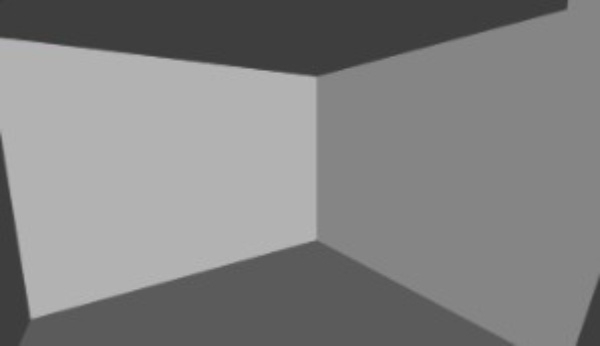}
  \caption{Room layout}
  \label{fig:filtering-Layout}
\end{subfigure}%
\begin{subfigure}{0.5\columnwidth}
  \includegraphics[width=\linewidth]{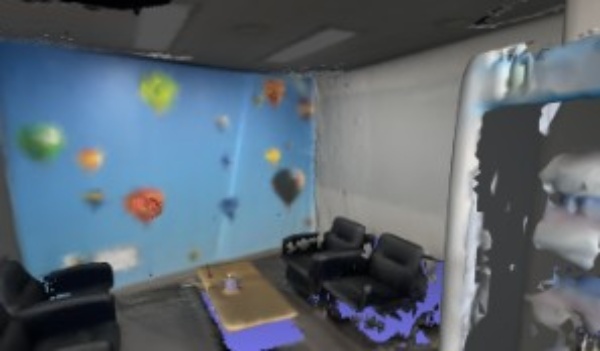}
  \caption{\(M_{\text{original}}\)}
  \label{fig:filtering-Orig}
\end{subfigure}\\
\begin{subfigure}{0.5\columnwidth}
  \includegraphics[width=\linewidth]{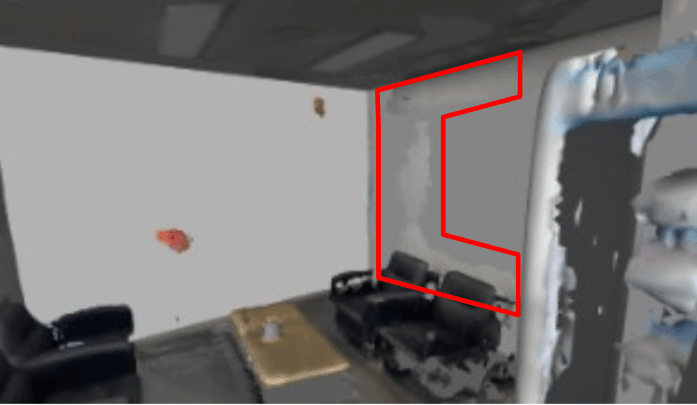}
  \caption{Room layout and \(M_{\text{original}}\) overlaid}
  \label{fig:nearbyFacesTobeFiltered}
\end{subfigure}%
\begin{subfigure}{0.5\columnwidth}
  \includegraphics[width=\linewidth]{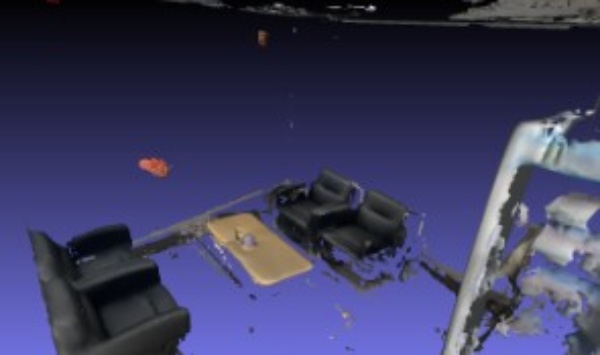}
  \caption{\(M_{\text{filtered}}\)}
  \label{fig:filtering-filtered}
\end{subfigure}\\
\begin{subfigure}{0.5\columnwidth}
  \includegraphics[width=\linewidth]{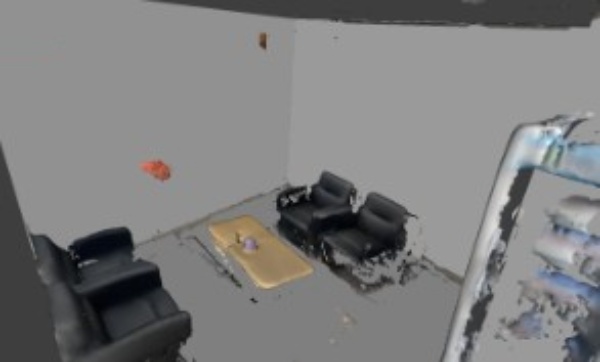}
  \caption{\(M_{\text{combined}}\)}
  \label{fig:filtering-combined1}
\end{subfigure}%
\begin{subfigure}{0.5\columnwidth}
  \includegraphics[width=\linewidth]{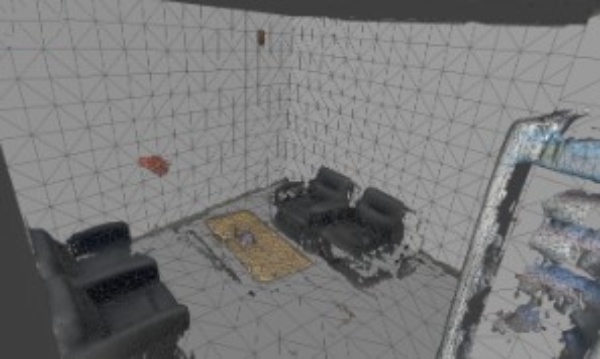}
  \caption{\(M_{\text{combined}}\) in wireframes}
  \label{fig:filtering-combined2}
\end{subfigure}
\caption{Detailed views of the Mesh Processing step, showing various stages from the initial room layout and \(M_{\text{original}}\) to the final processed mesh representation \(M_{\text{combined}}\).}
\label{fig:method:MeshProcessing}
\end{figure}

\noindent\cref{fig:method:MeshProcessing} provides detailed insights into the Mesh Processing step (Section 3.4). Starting with the room layout (\cref{fig:filtering-Layout}) and the original mesh (\cref{fig:filtering-Orig}), which is derived from the Capturing step (Section 3.3) as 
\(M_{\text{original}}\), the process continues with the mesh filtering process which eliminates mesh faces near the walls of the room layout structures (\cref{fig:nearbyFacesTobeFiltered}), resulting in the filtered mesh, \(M_{\text{filtered}}\) (\cref{fig:filtering-filtered}).

\noindent After the remeshing procedures are applied to the structural elements (vertical and horizontal walls), these remeshed structural meshes are merged with the filtered mesh. The result is the creation of the combined mesh (\cref{fig:filtering-combined1} and \cref{fig:filtering-combined2}), denoted as \(M_{\text{combined}}\), which serves as input for the subsequent Texturing step (Section 3.6).

\subsection{Further visualizations of the SampleMode of the Post Texturing step}

\begin{figure}[h]
 \centering
 \includegraphics[width=0.5\columnwidth]{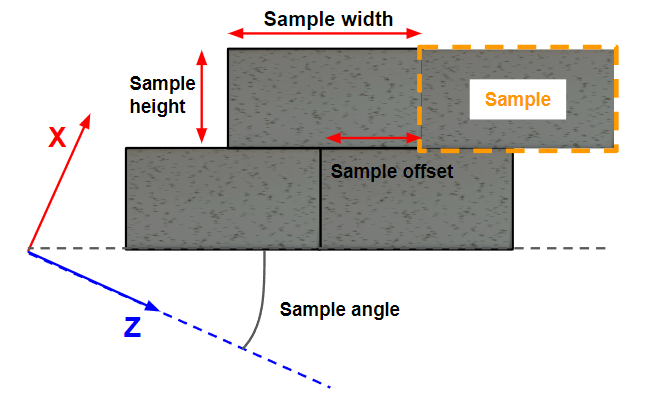}
 \caption{SampleMode parameters for the Post Texturing step.}
 \label{fig:method:sampleMode}
\end{figure}

\noindent In the Post Texturing step (Section 3.8)  of RoomRecon, texturing of walls can be done using the SampleMode. This approach allows the user to select either a custom or captured sample, which is then iteratively replicated to cover a specified plane of interest. Four adjustable parameters are provided as shown in \cref{fig:method:sampleMode}:
\begin{enumerate}
    \item Sample Width and Sample Height: Specify the dimensions of each sample.
    \item Sample Offset: Specifies the spacing between consecutive samples.
    \item Sample Angle: Allows rotation of the sample relative to the \(MBB_{align}\), the vector derived from the Mesh Processing step (Section 3.4) during the Minimum Bounding Box (MBB)~\cite{MBB} calculation.
\end{enumerate}

\begin{figure}[h]
\centering
\begin{subfigure}{0.4\columnwidth}
  \includegraphics[width=\linewidth]{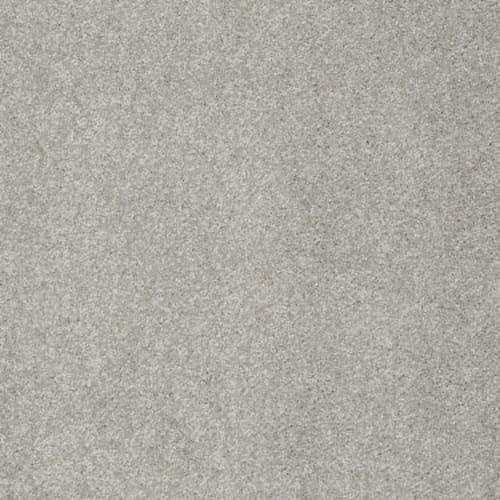}
  \caption{Floor sample used to generate floor texture}
  \label{fig:floorSample}
\end{subfigure}%
\hspace{10mm}
\begin{subfigure}{0.35\columnwidth}
  \includegraphics[width=\linewidth]{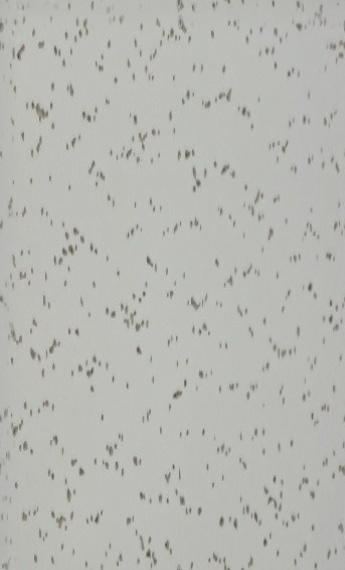}
  \caption{Ceiling sample used to generate ceiling texture}
  \label{fig:ceilingSample}
\end{subfigure}\\
\caption{Example samples for floor and ceiling texturing. Refer to \cref{fig:sampleModeFloor} and \cref{fig:sampleModeCeiling} for the generated textures.}
\label{fig:method:samplesForSampleMode}
\end{figure}

\begin{figure}[h]
 \centering
 \includegraphics[width=0.8\columnwidth]{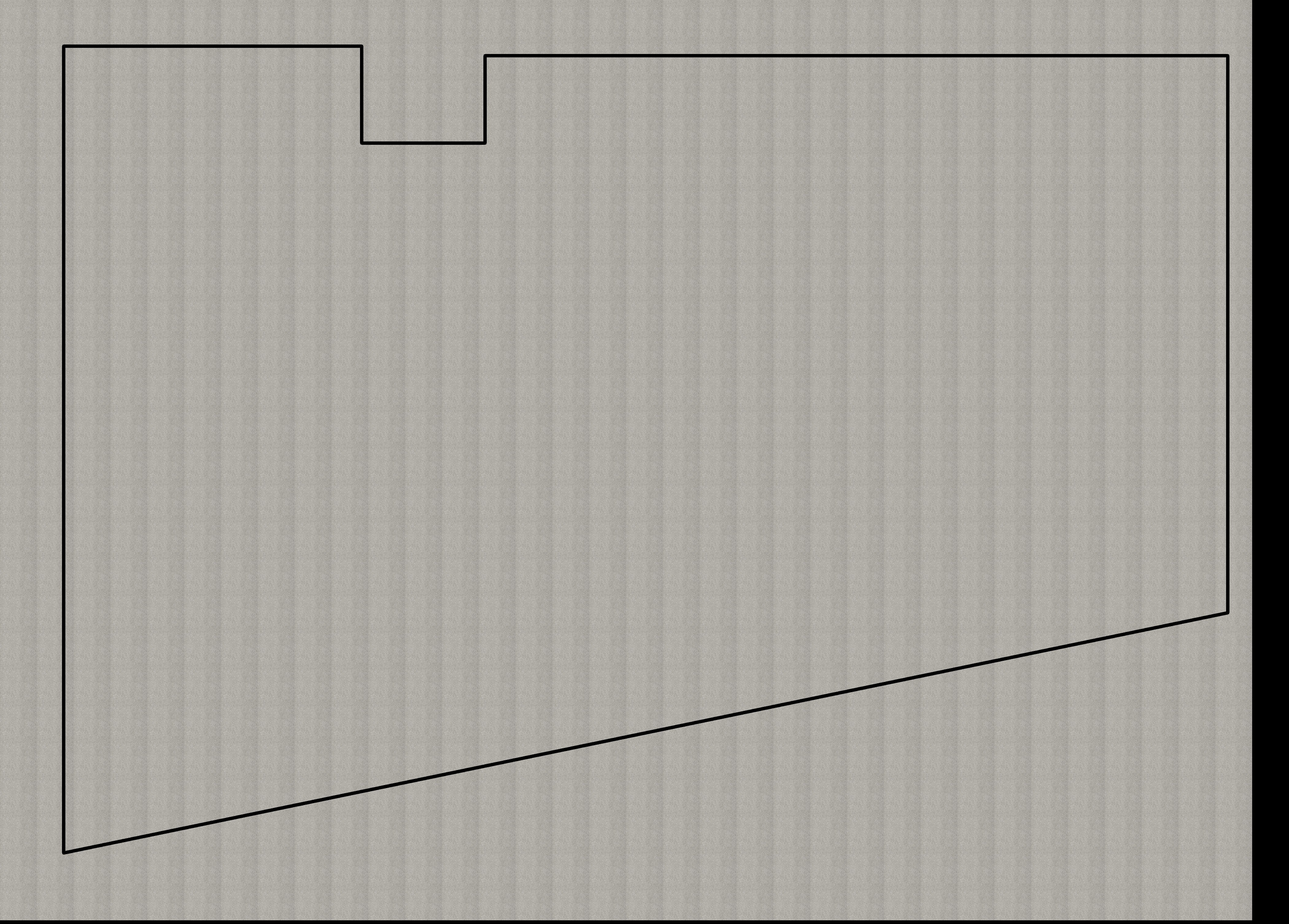}
 \caption{Example of a floor texture generated from the sample depicted in \cref{fig:floorSample}. The layout of the room is delineated by solid black lines.}
 \label{fig:sampleModeFloor}
\end{figure}

\begin{figure}[h]
 \centering
 \includegraphics[width=0.8\columnwidth]{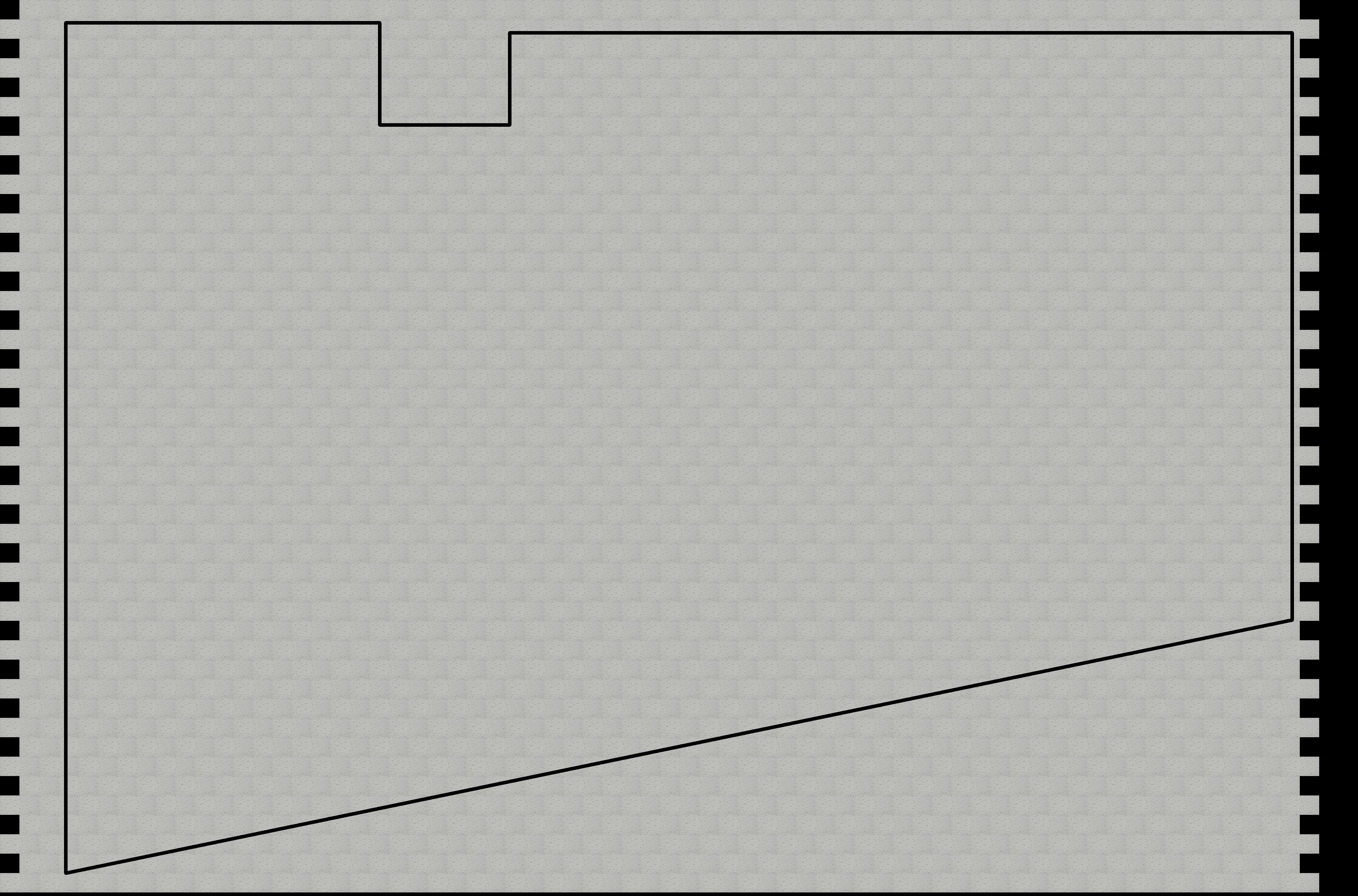}
 \caption{Example of a ceiling texture generated from the sample depicted in \cref{fig:ceilingSample}. The layout of the room is delineated by solid black lines.}
 \label{fig:sampleModeCeiling}
\end{figure}

Using samples shown in \cref{fig:floorSample} and \cref{fig:ceilingSample}, floor and ceiling textures can be created using the SampleMode as shown in \cref{fig:sampleModeFloor} and \cref{fig:sampleModeCeiling}, respectively.

\subsection{Extension of RoomRecon to MultiRoom}
\begin{figure}[h!]
 \centering
 \includegraphics[width=0.8\columnwidth]{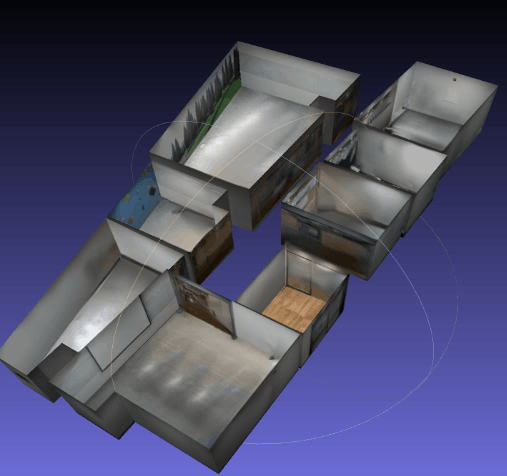}
 \caption{Extension of RoomRecon to MultiRoom.}
 \label{fig:multiRoom}
\end{figure}

In this paper, our primary goal is to introduce and evaluate the RoomRecon pipeline in the context of single-room reconstruction. However, the versatility of RoomRecon extends seamlessly to multi-room environments. By using RoomRecon to reconstruct rooms one by one, we can construct comprehensive multi-room environments that span entire floors, such as office spaces or apartments.

\noindent The RoomRecon pipeline, when applied to a multi-room environment, produces the result shown in \cref{fig:multiRoom}. This extension is straightforward, provided that the respective room layouts and camera images (specifically, camera extrinsics) adhere to a consistent world coordinate system.

\noindent This extension greatly expands the scope of the application by facilitating the creation of realistic and fully textured interiors over large areas. Its impact spans multiple domains, including augmented reality (AR), virtual reality (VR), robotics, retail and real estate.

\subsection{Further visualizations of the Texturing step}

\begin{figure}[h!]
 \centering
 \includegraphics[width=0.8\columnwidth]{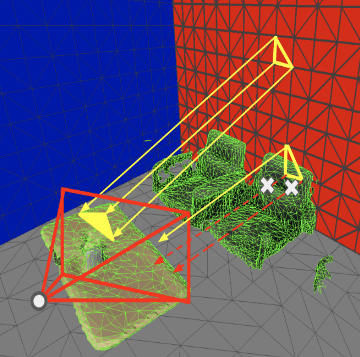}
 \caption{Occlusion test used in texturing methods~\cite{SEAMLESS, MVSTexturing} for an image and a face pair.}
 \label{fig:rayHitTest}
\end{figure}

In \cref{fig:rayHitTest}, we illustrate the occlusion test, a crucial component of assignment-based texturing methods~\cite{SEAMLESS, MVSTexturing}. This test is crucial for filtering candidate texture images for each mesh face during the Texturing step (Section 3.6). For a given image-face pair, successfully projecting three vertices of a face onto an image designates it as a candidate texture image for that mesh face. Conversely, if any ray connecting each vertex to a camera center is obstructed by other parts of the mesh, the image is eliminated as a candidate.

\noindent MVSTex~\cite{MVSTexturing} and SeamlessTex~\cite{SEAMLESS} employ additional refinement techniques, including back-face culling and angle criteria, to enhance the selection of candidate images. Specifically, the angle between a ray from the face center to the camera center, and the face normal should not exceed a certain angular threshold - e.g., 45 degrees. The face-image qualities are then stored in a sparse two-dimensional table, with rows representing images and columns representing faces. This table stores unary costs that represent the scores of the face-image pairs. During the Markov Random Field (MRF) optimization process for texture mapping, this table is used to determine the optimal assignment of texture images to mesh faces. In addition, pairwise costs are included in the MRF optimization to enforce the use of the same texture image for neighboring faces.

\noindent In our problem setting of creating a complete and high-fidelity 3D textured room layout, several technological and environmental challenges arise during the Texturing step (Section 3.6) that utilizes the assignment-based texturing methods~\cite{SEAMLESS, MVSTexturing}. These problems are grouped into eight distinct categories (A-H) as visualized in \cref{fig:typicalTexturingProblem}. 

\textbf{A}: Geometry and camera misalignments often occur due to incomplete and inaccurate meshes, as well as camera pose drift in VIO (Visual-Inertial Odometry) pose tracking systems~\cite{ARKIT}. An example of this is shown in Wall 6, where the chair textures invade the wall texture due to misaligned camera poses and incomplete chair meshes. Although the intent of the occlusion test (\cref{fig:rayHitTest}) is to exclude this image from texturing that specific part of the plane, the incomplete chair meshes result in its inclusion during the Texturing process~\cite{SEAMLESS, MVSTexturing}. Wall 4 also exhibits this issue, characterized by the texture of the refrigerator intruding into its surface texture.

\textbf{B}: Texture seams are caused by camera pose drift. Assignment-based texturing methods use multiple images to texture a single planar structure. If these images have misaligned poses, noticeable texture seams result. See Wall 6 in \cref{fig:typicalTexturingProblem}. 

\textbf{C}: Uncaptured areas and \textbf{D}: Occlusion present similar challenges, but from different sources. Uncaptured areas, such as those on Wall 1 and Wall 2, occur when users fail to capture certain portions of a wall. Conversely, occlusion problems occur when objects obstruct planar surfaces (Wall 3, Wall 4, Wall5, Wall 6), particularly floors due to many furniture on them.

\textbf{E}: The blending of object shadows onto wall textures, as shown in Wall 2, is an issue that is influenced by object placement and lighting conditions.

\textbf{F}: The thin-object problem occurs when slender structures are not fully captured during the initial Capturing step (Section 3.3), often due to depth sensor resolution limitations. In such cases, the mesh of thin objects is incomplete, leading to an erroneous result in the occlusion test (\cref{fig:rayHitTest}). Note that Wall 1 has a texture of clothes hanger on its surface texture.

\textbf{G} and \textbf{H}: Reflective and transparent surfaces introduce additional complexities. For example, reflective surfaces such as mirrors and whiteboards produce unwanted reflections, while transparent walls, such as glass walls, lead to inconsistent viewpoints (e.g., inconsistent views of outdoor scenes) from images used to texture a planar wall. Look at Wall 5 and 6 for these problems.

\noindent Ceilings and floors, which typically consist of repeating patterns, are particularly susceptible to camera pose drift. Even small discrepancies caused by pose drift result in conspicuous misaligned lines that significantly compromise visual aesthetics.

\subsection{Further visualization of inpainting and stable-diffusion}

\begin{figure}[h!]
  \centering
  \includegraphics[width=0.8\columnwidth]{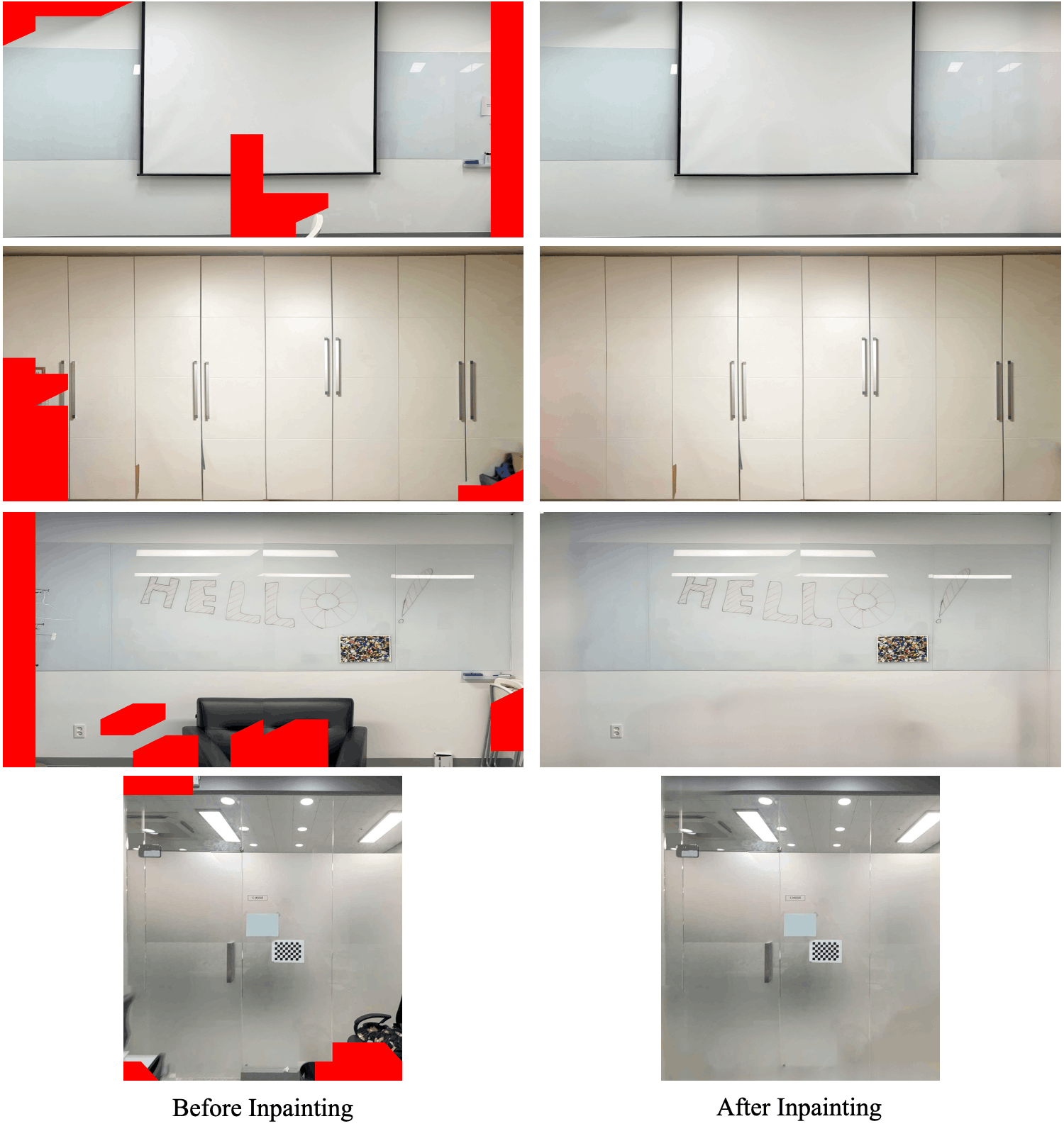}
  \caption{Examples of plane images before and after inpainting: We use ZIT~\cite{ZIT} as our inpainting model. Red masks are untextured areas caused by occlusion by furniture.
  }
\label{fig:inpainting}
\end{figure}

\begin{figure}[h!]
  \centering
  \includegraphics[width=0.6\columnwidth]{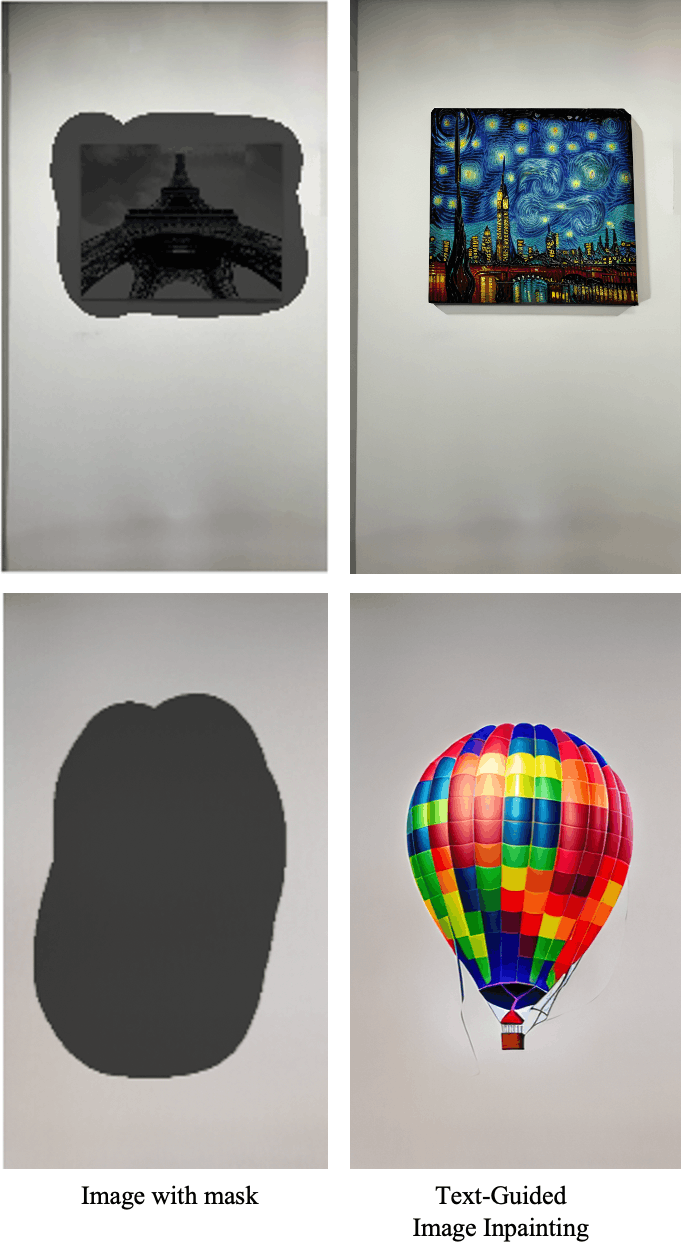}
  \caption{Examples of the Post Texturing (ImageMode) results using the stable diffusion model (PowerPaint~\cite{PowerPaint}), which is a server-based service on RoomRecon. In the first example (the first row), we input the text \textit{"A painting, hanging on the wall, portrays a city at night in the style of Van Gogh"}. In the second example (the second row), the input text is \textit{"a flying balloon"}.}
\label{fig:SD}
\end{figure}

\begin{figure*}[h!]
  \centering
  \includegraphics[width=1.0\linewidth]{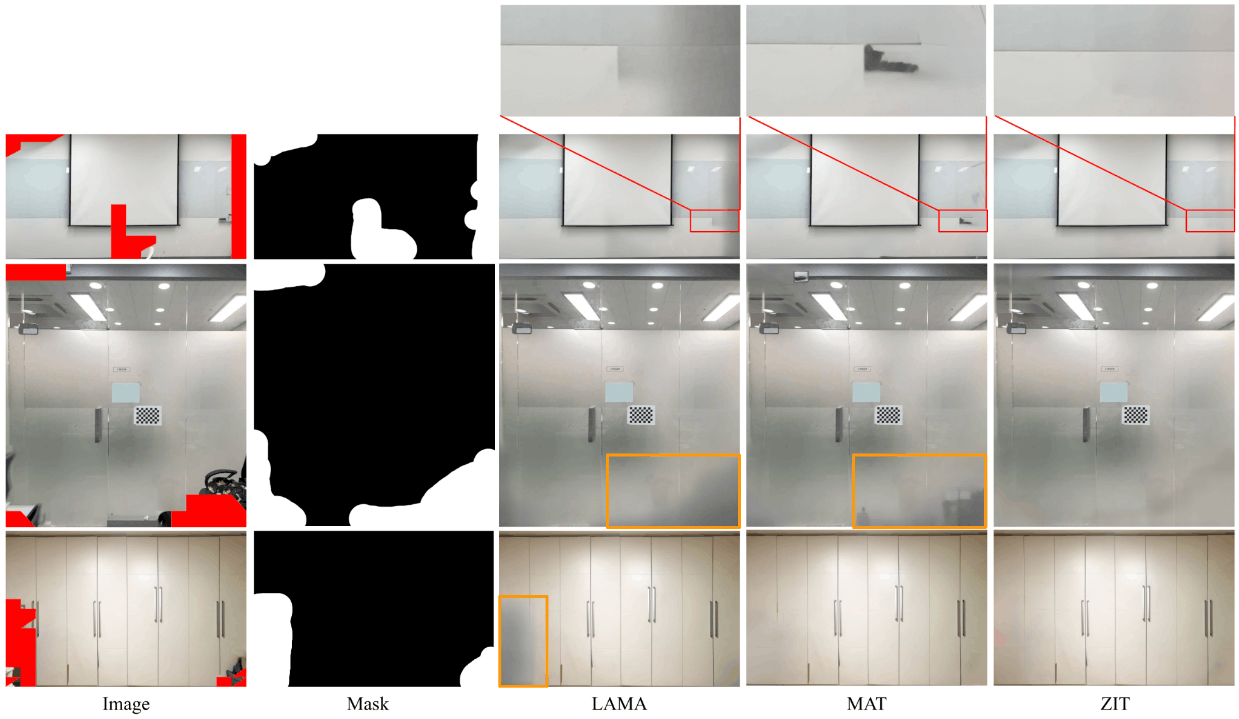}
  \caption{Visualisation of images inpainted by various methods, including LAMA~\cite{LamaInpainting}, MAT~\cite{MAT} and ZIT~\cite{ZIT}. Red areas indicate untextured areas caused by occlusion of furniture and inpainted using the inpainting methods.}
\label{fig:inpainting_comparison}
\end{figure*}

In \cref{fig:inpainting}, we present examples of images before and after inpainting with ZIT~\cite{ZIT}, demonstrating the effectiveness of inpainting in room layout reconstruction applications. While any inpainting method could be used for the Post-Texturing (Sec. 3.8), we specifically chose ZIT for the following reasons. ZIT uses the Transformer Structure Restoration module to restore missing structures such as edges and lines. This is particularly useful for interior scenes, which are constituted by a lot of boundary patterns (e.g. doors, windows and paintings). Furthermore, in the qualitative comparison, ZIT proves its suitability for the task of seamlessly filling in masked areas, as shown in \cref{fig:inpainting_comparison}. Taking the red box in the first row of \cref{fig:inpainting_comparison} as an example, LAMA fails to recover the different colours of the board and wall area. Also, MAT tends to replace the masked area with another object. In contrast, ZIT effectively produces a high-quality image, successfully reconstructing the masked areas with the matching colours of different objects. It also creates a crisp boundary between them.
Moreover, ZIT shows benefits of reducing blurry prediction often produced by LAMA and MAT (orange boxes). The final reason for choosing ZIT is the active update from the author. Recently, an extended version of ZIT (ZIT++~\cite{ZIT++}) is released, providing the improvement in both performance and run time. A head-to-head quantitative comparison between LAMA, MAT, and ZIT as well as information of ZIT++ can be found in the paper of ZIT++~\cite{ZIT++}. It is worth noting that other inpainting methods can be easily integrated in RoomRecon, based on the user’s requirements. For example, LAMA inpainting is faster than ZIT, reducing waiting time for users, but yields lower inpainting quality. \\

\noindent Furthermore, \cref{fig:SD} demonstrates the application of the Stable Diffusion model~\cite{PowerPaint} in image processing, facilitating the customization and personalization of wall textures to meet diverse user preferences.

\begin{figure*}[h]
  \centering
  \includegraphics[width=1.0\linewidth]{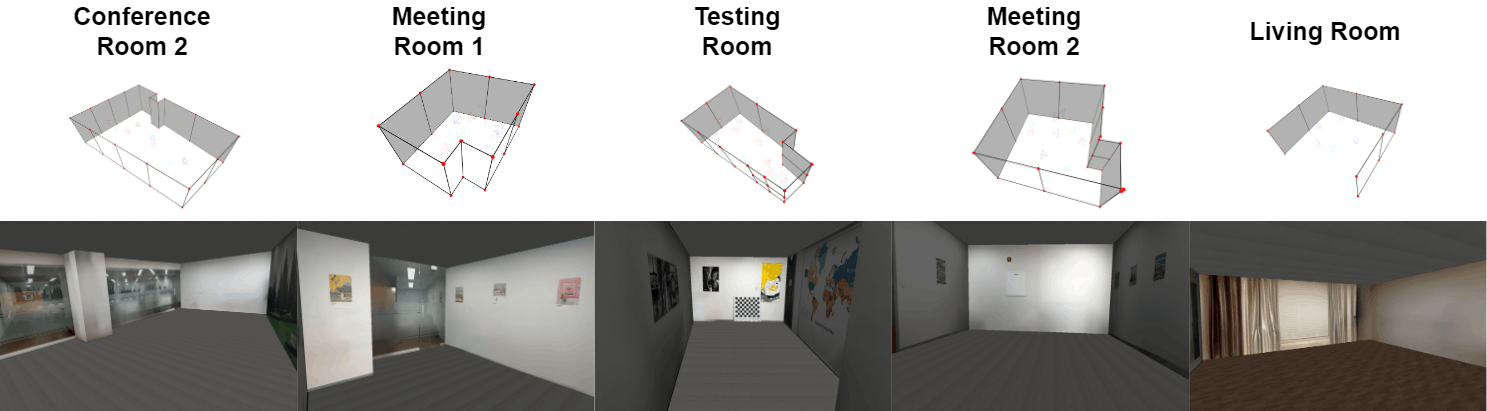}
  \caption{Additional views of five rooms from the paper.}
\label{fig:additionalPerspective}
\end{figure*}

\begin{figure*}[h]
 \centering
 \includegraphics[width=1.0\textwidth]{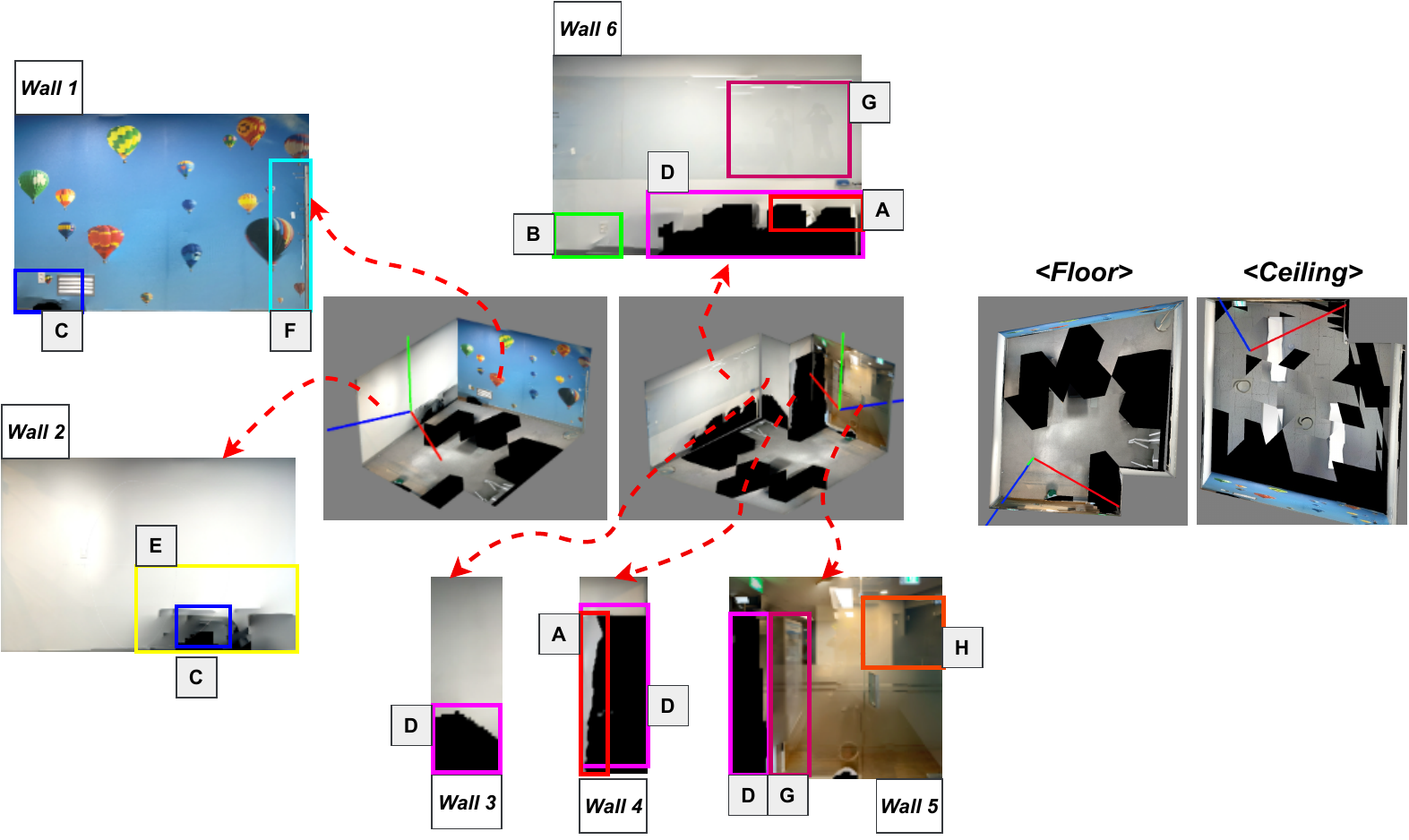}
 \caption{Texturing problems that occur in a room - A: Geometry and camera misalignment, B: Texture seams, C: Uncaptured areas, D: Occlusion, E: Object shadows, F: Thin objects, G: Reflection, H: Different views.}
 \label{fig:typicalTexturingProblem}
\end{figure*}

\begin{figure*}[ht]
  \centering
  \includegraphics[width=1.0\linewidth]{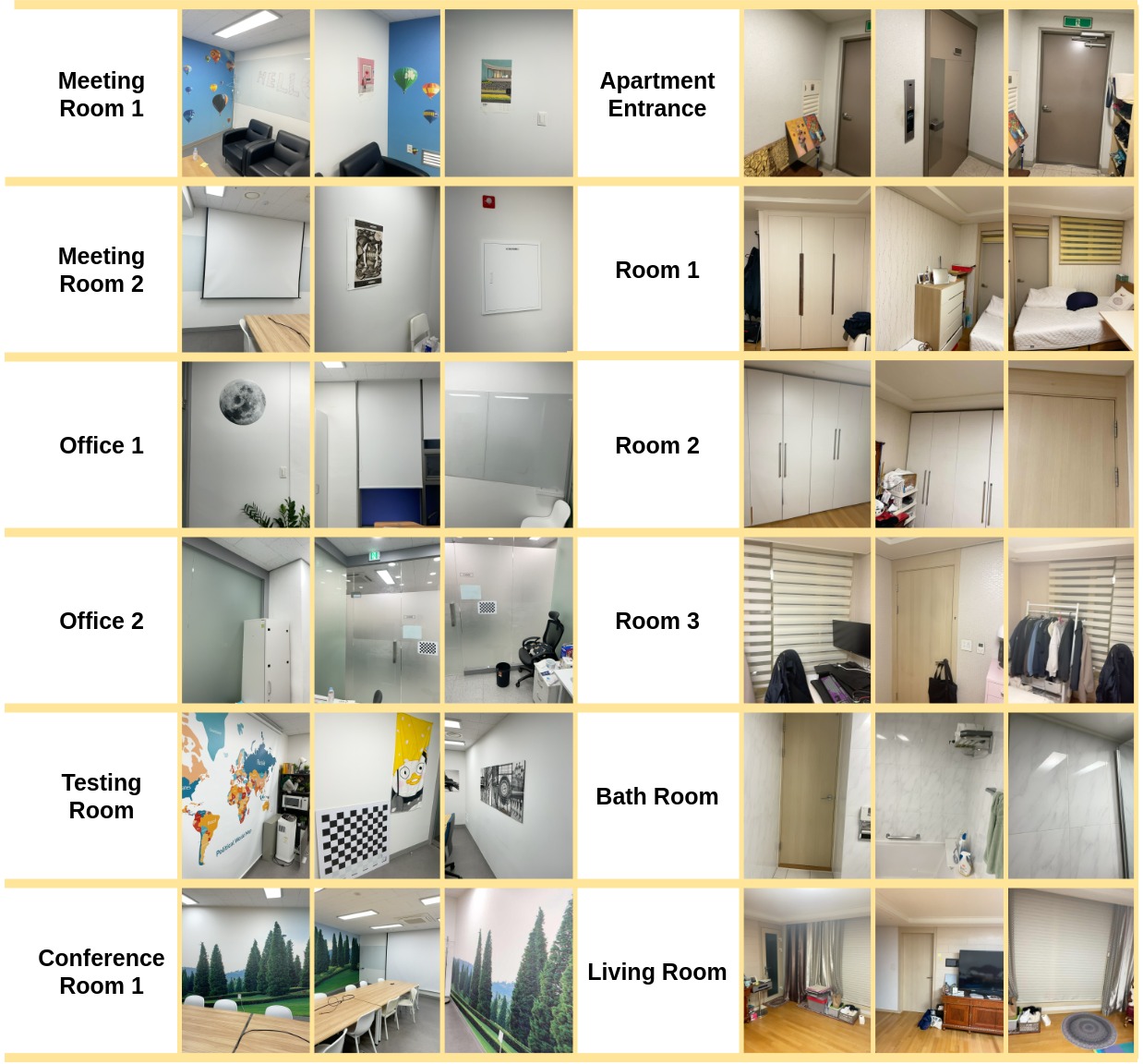}
  \caption{Representative samples of the ground truth images captured for each room.}
\label{fig:gtsamples}
\end{figure*}

\subsection{Ground truth image samples for local texture quality assessment}
For the local texture quality assessment (Sec. 4.2), the ground truth images were captured of areas with significant detail while avoiding regions with uniform textures such as plain white walls. We focused on areas such as door and window frames, wall paintings, and edges, where texture seams are highly noticeable and can lead to unsatisfactory results. This approach allowed us to effectively compare different texturing methods in these critical regions. \cref{fig:gtsamples} shows three representative samples of the ground truth images for each room. It can be seen that our objective was to capture areas with a high degree of detail with many lines/edges and colors.

\subsection{Further visualization of textured rooms}

\cref{fig:additionalPerspective} presents additional views of 3D textured models for five rooms discussed in the main paper (Fig. 9 of the paper), offering a comprehensive depiction of each textured room.

\begin{figure*}[ht]
  \centering
  \includegraphics[width=1.0\linewidth]{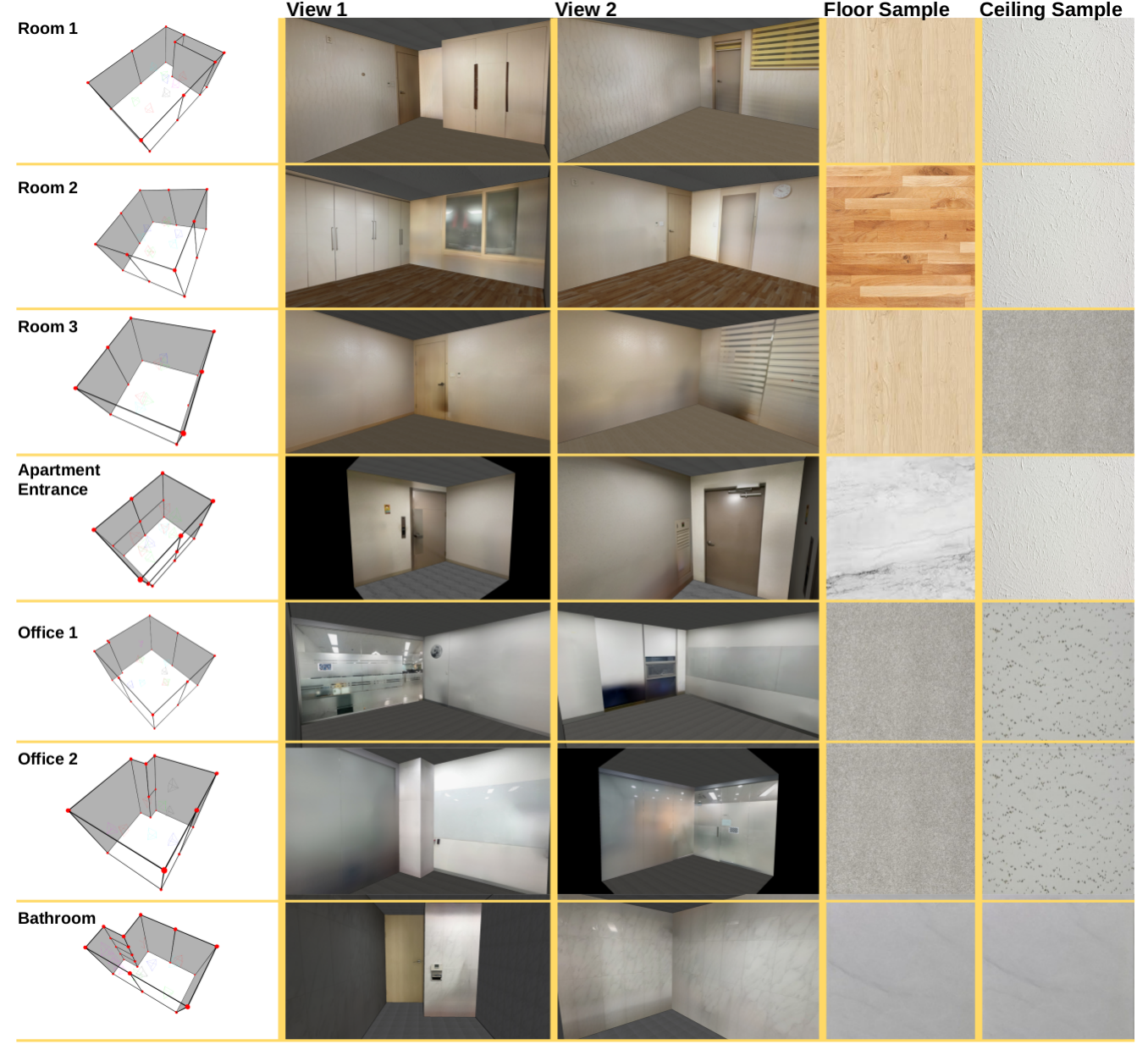}
  \caption{Visualization of textured rooms: The left column shows the room layouts, while the 2nd and 3rd columns show textured room layouts seen from two different perspectives. The 4th and 5th columns show samples used for the generation of floor and ceiling textures.
  }
\label{fig:results}
\end{figure*}

\noindent Next, \cref{fig:results} illustrates textured room layouts for 7 different rooms. The left column shows the room layouts, while columns two and three show textured rooms, viewed from two different perspectives. Columns four and five show samples used for floor and ceiling textures, created using the SampleMode of the Post Texturing step (Section 3.8).

\begin{figure*}[ht]
  \centering
  \includegraphics[width=1.0\linewidth]{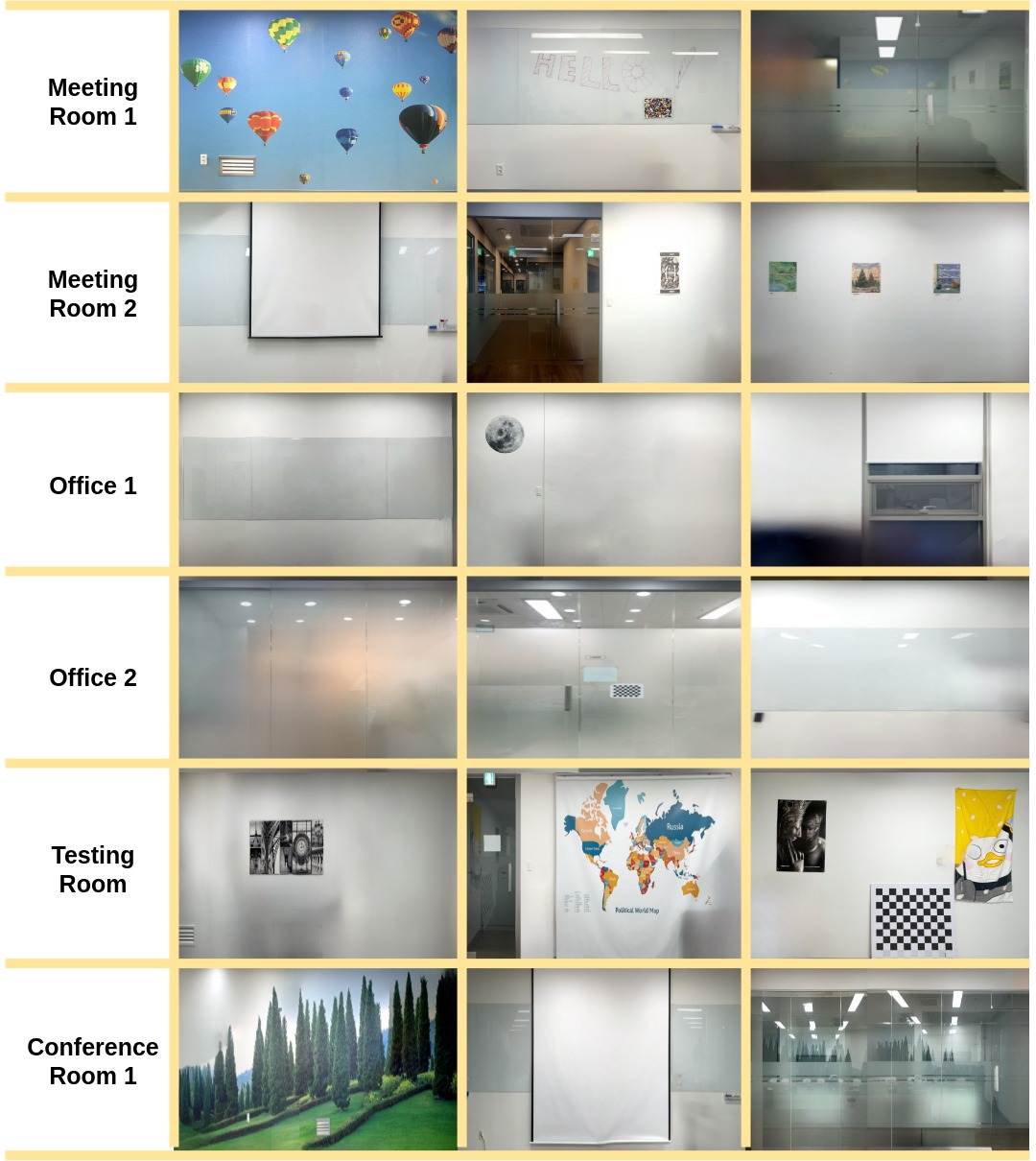}
  \caption{Additional visualizations of the textured walls for each room after the inpainting process - 1/2.}
\label{fig:add1}
\end{figure*}

\begin{figure*}[ht]
  \centering
  \includegraphics[width=1.0\linewidth]{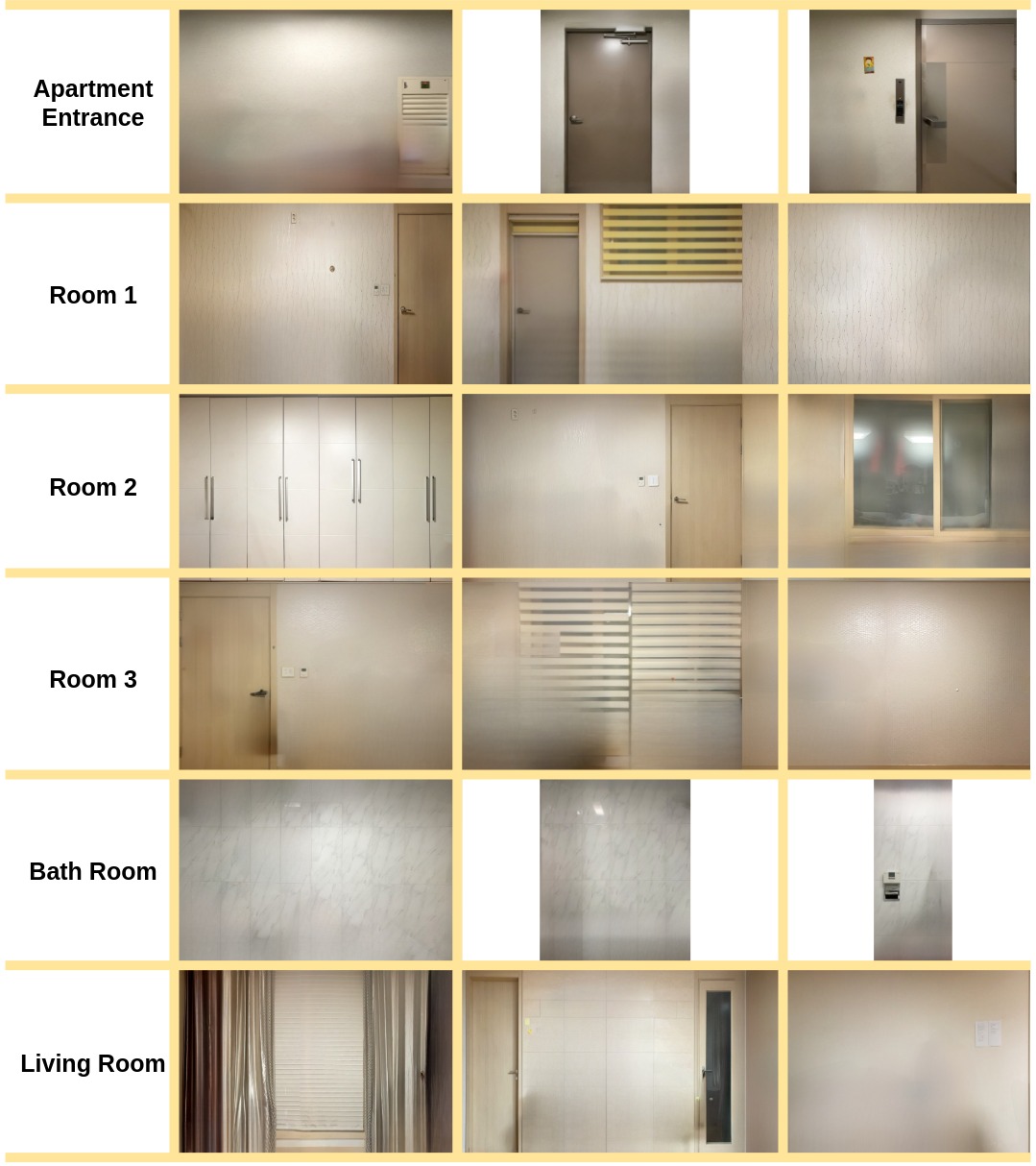}
  \caption{Additional visualizations of the textured walls for each room after the inpainting process - 2/2.}
\label{fig:add2}
\end{figure*}

\noindent Finally, \cref{fig:add1} and \cref{fig:add2} display three textured walls from each room, illustrating the high texturing quality achievable with the introduced pipeline.

\clearpage
\onecolumn

\section{Algorithms}
In this section, we present the pseudocode for algorithms discussed in the main paper. Initially, \cref{alg:plane2image} details the Plane2Image rendering process, as described in Section 3.7. Following this, \cref{alg:setupplane2image} elaborates on the computation of the perspective and view matrices essential for the rendering process. Next, \cref{alg:divideAndConquer} introduces the "divide" component of our Divide-And-Conquer scheme, which is covered in Section 3.5. This component focuses on determining optimal camera positions for capturing texture images of sub(planes).

\subsection{Plane2Image algorithm}
\begin{algorithm*}
\caption{Plane2Image algorithm}
\label{alg:plane2image}
\begin{algorithmic}[1]
  \Function{PLANE2IMAGE}{$planeIdx, isVertical, isFloor$}
    \If{$isVertical$} \Comment{Check if it is a vertical plane}
        \State $vertiInfo \gets \text{planeInfo}[planeIdx]$
        \State $planeNormal \gets \text{vertiInfo.planeNormal}$
        \State $planeCenter \gets \text{vertiInfo.planeCenter}$
        \State $planeWidth \gets \text{vertiInfo.planeWidth}$
        \State $planeHeight \gets \text{vertiInfo.planeHeight}$
    \Else
        \State $mbbCenter2D \gets \textbf{\(MBB_{center}\)}$
        \State $planeWidth \gets \textbf{\(MBB_{width}\)}$
        \State $planeHeight \gets \textbf{\(MBB_{height}\)}$
        \If{$isFloor$}
            \State $planeNormal \gets (0, 1, 0)$
            \State $planeCenter \gets (mbbCenter2D.x, \text{floorHeight}, \dots$ 
            \State $                   \phantom{mbbCenter2D.x, }mbbCenter2D.y)$
        \Else
            \State $planeNormal \gets (0, -1, 0)$
            \State $planeCenter \gets (mbbCenter2D.x, \text{ceilingHeight}, \dots$ 
            \State $                   \phantom{mbbCenter2D.x, }mbbCenter2D.y)$
        \EndIf
    \EndIf
    \State $viewMatrix, projectionMatrix \gets \text{SETUPCAM}(planeCenter, planeNormal, planeWidth, planeHeight, isFloor)$

    \State $RENDER(viewMatrix, projectionMatrix)$
  \EndFunction
\end{algorithmic}
\end{algorithm*}

\begin{algorithm*}[h!]
\caption{Set up a camera for plane rendering for the Plane2Image operation}
\label{alg:setupplane2image}
\begin{algorithmic}[1]
    \Function{SetupCam}{$center, normal, width, height, isVertical, isFloor$}
        \State $perMeterPixelCount \gets 500$ \Comment{Defines the resolution in pixels per meter.}
        \State $lookAtVector \gets center$
        \State $cameraPosition \gets center + normal \times height$
        \State $upVector \gets (0, 1, 0)$ \Comment{Default up vector}
        \If{Not $isVertical$}
            \If{$isFloor$}
                \State $upVector \gets -\textbf{MBB}_{\text{align}}$
            \Else
                \State $upVector \gets \textbf{MBB}_{\text{align}}$
            \EndIf
        \EndIf
        \State $viewMatrix \gets \text{calculateViewMatrix}(cameraPosition, lookAtVector, upVector)$
        \State $planePixelWidth \gets \text{int}(perMeterPixelCount \times width)$ 
        \State $planePixelHeight \gets \text{int}(perMeterPixelCount \times height)$ 
        \State $near \gets 0.001f$ \Comment{Near clipping plane for rendering}
        \State $far \gets 100.0f$ \Comment{Far clipping plane for rendering}
        \State $fx \gets planePixelHeight / 2.0$
        \State $fy \gets planePixelHeight / 2.0$
        \State $cx \gets planePixelWidth / 2.0$
        \State $cy \gets planePixelHeight / 2.0$
        \State $imageWidth \gets planePixelWidth$
        \State $imageHeight \gets planePixelHeight$
        
        \State $projectionMatrix \gets \text{calculateProjectionMatrix}(planePixelWidth, planePixelHeight, fx, fy, cx, cy, near, far)$
        
        \State \Return $viewMatrix, projectionMatrix$
    \EndFunction
\end{algorithmic}
\end{algorithm*}

\subsection{Divide-And-Conquer algorithm}
\begin{algorithm*}[h!]
\caption{Divide-And-Conquer algorithm for Texture Image Acquisition}
\label{alg:divideAndConquer}
\begin{algorithmic}[1]
  \Function{DivideAndConquer}{$planeIdx, walls$}
    \State $info \gets walls[planeIdx]$
    \State $normal \gets info.planeNormal$ \Comment{Plane normal}
    \State $center \gets info.planeCenter$ \Comment{Plane center}
    \State $width \gets info.planeWidth$ \Comment{Plane width}
    \State $height \gets info.planeHeight$ \Comment{Plane height}
    \State $fov \gets \text{Camera's field of view}$ 
    \State $optPosition \gets center + normal \times \text{optimal distance}$
    \State $canCapture \gets \Call{CanCapture}{width, height, fov, optPosition}$
    \If{$canCapture$ \textbf{and} \Call{WithinWalls}{optPosition, walls} \textbf{and} $\text{optimal distance} \leq 5$}
        \State \textbf{return} $optPosition$
    \Else
        \State $subPlanes \gets \Call{DividePlane}{width, height, 2}$
        \For{$sp$ in $subPlanes$}
            \State \textbf{return} $\Call{DivideAndConquer}{sp, walls}$ \Comment{Recursion until the conditions are met}
        \EndFor
    \EndIf
  \EndFunction
  \vspace{5mm}
  
  \Function{CanCapture}{$w, h, fov, pos$}
    \State \textbf{return} $(w \leq fov.width) \text{ and } (h \leq fov.height)$
  \EndFunction
  \vspace{5mm}
  
  \Function{WithinWalls}{$pos, walls$}
    \State \textbf{return} $\text{True if } pos \text{ is inside the polygon formed by } walls$
  \EndFunction
  \vspace{5mm}
  
  \Function{DividePlane}{$w, h, factor$}
    \State $newW \gets w / factor$
    \State $newH \gets h / factor$
    \State \textbf{return} $[(newW, newH), (newW, newH)]$ \Comment{Splitted to two subplanes}
  \EndFunction

\end{algorithmic}
\end{algorithm*}
\section{Quantitative statistics of experimental datasets}
\label{Supple:table}
In this section, we present the comprehensive results of the experiments described in the main paper. It is important to note that the numbers given in the main paper represent averages, which are also provided at the bottom of each table for clarity and reference.

\noindent \cref{tab:supp:iosProcessTime} presents the processing time measurements for each step of the RoomRecon pipeline on an iOS device, specifically an iPhone 12 Pro with an A14 Bionic chip, excluding texturing-related processes. The time measurements for the texturing-related processes are detailed in \cref{tab:supp:texturingTimeBeforeAfter}.

\noindent \cref{tab:supp:iosProcessTime} outlines the following processes:
\begin{enumerate}
    \item Layout Parsing: This step involves acquiring layout information, including the normals, centers, and four vertices of the vertical walls, as well as the floor and ceiling heights.
    \item Loop Check and Form Floor and Ceiling: Loop Check step verifies whether the vertical walls form a closed loop, essential for subsequent mesh processing steps. Using this information, floor and ceiling meshes are generated using the Constrained Delaunay Triangulation (CDT) method~\cite{CDT} or the Minimum Bounding Box (MBB)~\cite{MBB} as described in Section 3.4 of the main paper.
    \item Mesh Processing (Section 3.4): This process includes mesh filtering, remeshing, and combining to produce the combined mesh (\(M_{\text{combined}}\)).
    \item Plane2Image (Section 3.7): This step involves rendering the textured meshes using our Plane2Image module.
    \item Displaying Results: Finally, the textured meshes (\(M_{\text{textured}}\)) are added to the scene for visualization and further Post Texturing processes (Section 3.8).
\end{enumerate}

\noindent \cref{tab:supp:texturingTimeBeforeAfter} presents a comparative analysis of the computational time required for texturing-related processes on an iOS device, an iPhone 12 Pro with an A14 Bionic chip, before and after the implementation of our Divide-And-Conquer scheme. The texturing method used in this experiment, MVSTex~\cite{MVSTexturing}, initially involved calculating the blurriness of each image to sample sharp images from \(C_{original}\), represented as Blur Calculation Time. However, our Divide-And-Conquer approach eliminates the need to compute blurriness because all captured images (\(C_{recaptured}\)) are consistently sharp. The computational overhead for the "divide" operation (Division Calculation Time) is minimal. By using a significantly reduced number of images (from 258.85 to 11.39), the texturing process using \(C_{recaptured}\) (Our-MVSTex) is accelerated, now completing in only 4.72 seconds on an iOS device.

\noindent \cref{tab:supp::blurAnalysis} provides a quantitative analysis of the number of frames (Total Frames, Sampled Frames, Recaptured Frames, Ground truth (GT) Frames) and their blur measures. Our recaptured images show sharper characteristics than other image groups (Total Frames, Sampled Frames) in all rooms. Blur metric used in this experiment is from the work of Crete et al.~\cite{Blurriness}.

\noindent \cref{tab:time_metrics} presents a texturing time analysis of all texturing methods~\cite{SEAMLESS, MVSTexturing, PlaneOpt, COLORMAPOPTIM} used in the experiment on a Linux computer equipped with an AMD Ryzen 9 5900x 12-core × 24 processor. The GPU specifications of the machine are not reported because none of the texturing methods involve GPU operations.

\begin{table*}[h]
\centering
\resizebox{\textwidth}{!}{%
\begin{tabular}{c|cccccc|c}
\toprule
 & \textbf{Layout} & \textbf{Loop} & \textbf{Form} & \textbf{Mesh} & \textbf{} & \textbf{Displaying} & \multicolumn{1}{c}{\textbf{}} \\
 & \textbf{Parsing} & \textbf{Check} & \textbf{Floor / Ceiling} & \textbf{Processing} & \textbf{Plane2Image} & \textbf{Results} & \multicolumn{1}{c}{\textbf{Total}} \\
 & \textbf{{[}ms{]}} & \textbf{{[}ms{]}} & \textbf{{[}ms{]}} & \textbf{{[}ms{]}} & \textbf{{[}ms{]}} & \textbf{{[}ms{]}} & \multicolumn{1}{c}{\textbf{{[}ms{]}}} \\ 
 \midrule
\textbf{Meeting Room 1} & 3.291 & 0.044 & 2.229 & 239.042 & 4291.335 & 476.295 & 5012.236 \\
\textbf{Meeting Room 2} & 5.082 & 0.028 & 2.195 & 344.215 & 4550.166 & 751.736 & 5653.422 \\
\textbf{Office 1} & 4.378 & 0.051 & 2.299 & 115.617 & 3798.089 & 382.080 & 4302.513 \\
\textbf{Office 2} & 5.342 & 0.049 & 2.264 & 265.978 & 4498.490 & 566.005 & 5338.128 \\
\textbf{Testing Room} & 3.506 & 0.029 & 2.525 & 339.635 & 5597.870 & 632.177 & 6575.742 \\
\textbf{Conference Room 1} & 3.823 & 0.016 & 2.133 & 219.718 & 3506.317 & 1212.163 & 4944.170 \\
\textbf{Conference Room 2} & 6.304 & 0.033 & 3.466 & 693.724 & 7042.335 & 1449.250 & 9195.112 \\
\textbf{Apartment Entrance} & 3.125 & 0.017 & 1.353 & 43.225 & 2643.013 & 276.860 & 2967.593 \\
\textbf{Room 1} & 3.412 & 0.115 & 2.202 & 197.832 & 4843.752 & 466.239 & 5513.551 \\
\textbf{Room 2} & 1.915 & 0.017 & 1.411 & 124.408 & 1690.656 & 410.992 & 2229.400 \\
\textbf{Room 3} & 2.120 & 0.017 & 1.419 & 114.148 & 3121.076 & 339.858 & 3578.637 \\
\textbf{Bath Room} & 3.527 & 0.026 & 2.167 & 78.829 & 3409.786 & 284.500 & 3778.835 \\
\textbf{Living Room} & 2.284 & 0.020 & 1.394 & 203.665 & 2063.789 & 613.244 & 2884.395 \\ 
\midrule
\textbf{Aggregated} & 3.701 & 0.036 & 2.081 & 229.234 & 3927.436 & 604.723 & 4767.210 \\
\textbf{Std Dev.} & \(\pm\)1.298 & \(\pm\)0.027 & \(\pm\)0.588 & \(\pm\)168.003 & \(\pm\)1455.371 & \(\pm\)354.626 & \(\pm\)1836.356 \\
\bottomrule
\end{tabular}%
}
\caption{Detailed timing analysis for each step of RoomRecon on iOS devices.}
\label{tab:supp:iosProcessTime}
\end{table*}

\begin{table*}[h]
\centering
\resizebox{\textwidth}{!}{%
\begin{tabular}{@{}c|cccc|cccc@{}}
\toprule
& \textbf{Number of} & \textbf{Blur Calculation} & \textbf{Number of} & \textbf{Texturing} & \textbf{Division Calculation} & \textbf{Number of} & \textbf{Texturing} \\
& \textbf{original frames (\(C_{orig.})\)} & \textbf{Time} & \textbf{selected frames} & \textbf{Time-Before} & \textbf{Time} & \textbf{recaptured frames (\(C_{recaptured})\)} & \textbf{Time-Our} \\
& \textbf{{[}-{]}} & \textbf{{[}ms{]}} & \textbf{{[}-{]}} & \textbf{{[}ms{]}} & \textbf{{[}ms{]}} & \textbf{{[}-{]}} & \textbf{{[}ms{]}} \\ 
\midrule
\textbf{Meeting Room 1} & 919 & 7420.991 & 405 & 23307.680 & 0.214 & 9 & 4911.648 \\
\textbf{Meeting Room 2} & 503 & 4038.300 & 233 & 22292.851 & 0.239 & 14 & 4960.514 \\
\textbf{Office 1} & 528 & 4342.993 & 230 & 13908.482 & 0.228 & 9 & 3179.689 \\
\textbf{Office 2} & 380 & 3144.345 & 179 & 21819.396 & 0.207 & 16 & 6263.140 \\
\textbf{Testing Room} & 1148 & 9530.300 & 471 & 34307.223 & 0.296 & 17 & 6654.108 \\
\textbf{Conference Room 1} & 779 & 6539.145 & 327 & 25441.619 & 0.259 & 8 & 4293.619 \\
\textbf{Conference Room 2} & 865 & 7317.129 & 370 & 34061.953 & 0.284 & 18 & 9489.955 \\
\textbf{Apartment Entrance} & 481 & 3816.633 & 200 & 7705.158 & 0.231 & 9 & 1928.829 \\
\textbf{Room 1} & 431 & 3306.680 & 185 & 13008.757 & 0.242 & 14 & 4440.505 \\
\textbf{Room 2} & 273 & 2091.293 & 129 & 7491.302 & 0.175 & 8 & 3414.745 \\
\textbf{Room 3} & 680 & 5267.885 & 290 & 19219.999 & 0.190 & 8 & 4186.139 \\
\textbf{Bath Room} & 282 & 2128.644 & 125 & 7481.924 & 0.228 & 10 & 1804.568 \\
\textbf{Living Room} & 530 & 4162.328 & 221 & 13170.632 & 0.188 & 8 & 5785.215 \\ 
\midrule
\textbf{Aggregated} & 599.923 & 4854.359 & 258.846 & 18708.998 & 0.229 & 11.385 & 4716.360 \\
\textbf{Std Dev.} & \(\pm\)262.900 & \(\pm\)2243.389 & \(\pm\)106.981 & \(\pm\)9253.167 & \(\pm\)0.036 & \(\pm\)3.820 & \(\pm\)2061.859 \\
\bottomrule
\end{tabular}%
}
\caption{Comparative analysis of texturing times before and after the introduction of \textbf{Divide And Conquer} on iOS devices.}
\label{tab:supp:texturingTimeBeforeAfter}
\end{table*}

\clearpage
\begin{table*}[h]
  \caption{Quantitative analysis of number of frames (Total Frames, Sampled Frames, Recaptured Frames, GT Frames) and blur metrics (Total Blur, Sampled Blur, Recaptured Blur) across rooms}
  \label{tab:supp::blurAnalysis}
  \centering
  \begin{tabular}{lccccccc}
    \toprule
    Room Type & \begin{tabular}[c]{@{}c@{}}Total\\Frames\end{tabular} & \begin{tabular}[c]{@{}c@{}}Sampled\\Frames\end{tabular} & \begin{tabular}[c]{@{}c@{}}Recaptured\\Frames\end{tabular} & \begin{tabular}[c]{@{}c@{}}GT\\Frames\end{tabular} & \begin{tabular}[c]{@{}c@{}}Total\\Blur $\downarrow$\end{tabular} & \begin{tabular}[c]{@{}c@{}}Sampled\\Blur $\downarrow$\end{tabular} & \begin{tabular}[c]{@{}c@{}}Recaptured\\Blur $\downarrow$\end{tabular} \\
    \midrule
    Meeting Room 1 & 919 & 405 & 9 & 20 & 0.38 & 0.34 & 0.23 \\
    Meeting Room 2 & 503 & 233 & 14 & 15 & 0.37 & 0.34 & 0.22 \\
    Office 1 & 528 & 230 & 9 & 14 & 0.37 & 0.35 & 0.24 \\
    Office 2 & 380 & 179 & 16 & 15 & 0.37 & 0.33 & 0.25 \\
    Testing Room & 1148 & 471 & 17 & 16 & 0.33 & 0.31 & 0.21 \\
    Conference Room 1 & 779 & 327 & 8 & 17 & 0.35 & 0.32 & 0.23 \\
    Conference Room 2 & 865 & 370 & 18 & 20 & 0.36 & 0.34 & 0.26 \\
    Apartment Entrance & 481 & 200 & 9 & 12 & 0.44 & 0.40 & 0.25 \\
    Room 1 & 431 & 185 & 14 & 13 & 0.50 & 0.46 & 0.27 \\
    Room 2 & 273 & 129 & 8 & 15 & 0.47 & 0.43 & 0.29 \\
    Room 3 & 680 & 290 & 8 & 10 & 0.49 & 0.46 & 0.26 \\
    Bath Room & 282 & 125 & 10 & 14 & 0.43 & 0.39 & 0.25 \\
    Living Room & 530 & 221 & 8 & 20 & 0.44 & 0.41 & 0.28 \\
    \midrule
    Aggregated & 599.92 & 258.85 & 11.38 & 15.46 & 0.41 & 0.37 & 0.25 \\
    Std Dev & \(\pm\)262.90 & \(\pm\)106.98 & \(\pm\)3.82 & \(\pm\)3.13 & \(\pm\)0.06 & \(\pm\)0.05 & \(\pm\)0.02 \\
    \bottomrule
  \end{tabular}
\end{table*}


\begin{table*}[htbp]
  \caption{Comparison of time metrics for different texture methods (performed on Linux machine).}
  \label{tab:time_metrics}
  \centering
  \begin{tabular}{lrrrrrr}
    \toprule
    Room Type & \begin{tabular}[c]{@{}r@{}}ColorMapOpt~\cite{COLORMAPOPTIM}\\(ms)\end{tabular} & \begin{tabular}[c]{@{}r@{}}PlaneOpt~\cite{PlaneOpt} \\(ms)\end{tabular} & \begin{tabular}[c]{@{}r@{}}MVSTex~\cite{MVSTexturing}\\(ms)\end{tabular} & \begin{tabular}[c]{@{}r@{}}SeamlessTex~\cite{SEAMLESS}\\(ms)\end{tabular} & \begin{tabular}[c]{@{}r@{}}Our-MVSTex\\(ms)\end{tabular} & \begin{tabular}[c]{@{}r@{}}Our-SeamlessTex\\(ms)\end{tabular} \\
    \midrule
    Meeting Room 1 & 115352 & 204867 & 12867 & 43502 & 3163 & 3912 \\
    Meeting Room 2 & 61712 & 347866 & 7692 & 25101 & 2555 & 3885 \\
    Office 1 & 70766 & 109676 & 8050 & 28356 & 2941 & 3862 \\
    Office 2 & 98909 & 197003 & 7218 & 20941 & 4526 & 5937 \\
    Testing Room & 137755 & 290077 & 7979 & 87402 & 4179 & 5963 \\
    Conference Room 1 & 62039 & 684336 & 8908 & 36407 & 2418 & 3293 \\
    Conference Room 2 & 140178 & 521584 & 10357 & 41069 & 5246 & 6665 \\
    Apartment Entrance & 50564 & 125640 & 7235 & 22719 & 1977 & 2989 \\
    Room 1 & 70610 & 104626 & 7413 & 21395 & 3793 & 5766 \\
    Room 2 & 27317 & 373976 & 7049 & 14887 & 2755 & 4044 \\
    Room 3 & 65555 & 211477 & 9191 & 34844 & 2832 & 3956 \\
    Bath Room & 36651 & 55988 & 5108 & 14156 & 1600 & 2535 \\
    Living Room & 83148 & 332202 & 11178 & 33278 & 3960 & 4972 \\
    \midrule
    Aggregated[s] & 78.50 & 273.79 & 8.48 & 32.62 & 3.23 & 4.44 \\
    Std Dev[s] & \(\pm\)35.48 & \(\pm\)179.96 & \(\pm\)2.03 & \(\pm\)18.93 & \(\pm\)1.05 & \(\pm\)1.29 \\
    \bottomrule
  \end{tabular}
\end{table*}

\end{document}